\documentclass[10pt,twocolumn,letterpaper]{article}
\usepackage{multirow}
\usepackage{booktabs}
\usepackage{xcolor}
\usepackage{colortbl}
\usepackage{graphicx}
\usepackage{caption}
\usepackage{amsmath}
\usepackage{amssymb}
\usepackage{bbm}
\usepackage{bbding}
\usepackage{pifont}

\usepackage{makecell}      % 表格内换行
\usepackage{graphicx}      % 插图
\usepackage{booktabs}      % \toprule 等
\usepackage{array}         % m{..} 列支持
\usepackage{adjustbox}     % 表格整体缩放
\usepackage{subcaption}

\usepackage{tabularx, makecell, booktabs, adjustbox, array, ragged2e} 
\usepackage[most]{tcolorbox}
\newcommand{\concept}{\textbf{\textit{Image-Conditioned Questioner}}}
\newtcolorbox{alprompt}[1]{
        boxrule = 1pt,
        fontupper = \small\tt,
        fonttitle = \bf\color{black},
        arc = 2pt,
        rounded corners,
        colframe = black,
        colbacktitle = white!97!yellow,
        colback = white!97!yellow,
        title = #1,
}
\usepackage[linesnumbered,ruled,vlined]{algorithm2e}
\SetAlgoNlRelativeSize{-1}  % 调整行号大小
\SetKwInput{KwInput}{Input}                % 输入
\SetKwInput{KwOutput}{Output}              % 输出
\SetKwComment{Comment}{$\triangleright$\ }{} 

\usepackage{cvpr}              % To produce the CAMERA-READY version
\newcommand{\method}{\textit{VisPlay}}

\definecolor{cvprblue}{rgb}{0.21,0.49,0.74}
\usepackage[pagebackref,breaklinks,colorlinks,allcolors=cvprblue]{hyperref}

\def\paperID{74
21} % *** Enter the Paper ID here
\def\confName{CVPR}
\def\confYear{2026}

\title{
  \raisebox{-1em}{\includegraphics[height=3.2em]{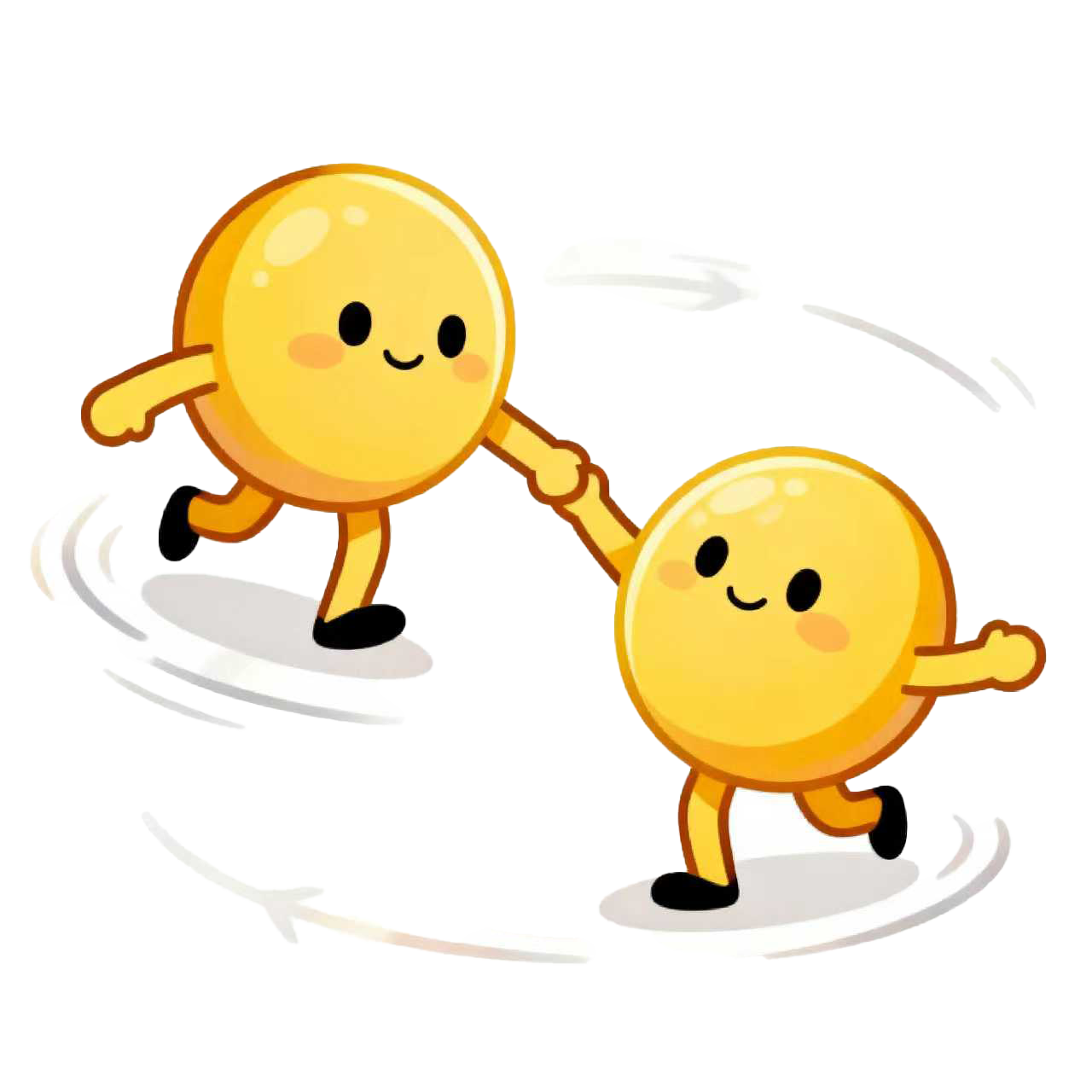}\hspace{0.5em}}
  \method: Self-Evolving Vision-Language Models from Images
}
\author{Yicheng He\textsuperscript{1}\thanks{Core Contribution.} \quad Chengsong Huang\textsuperscript{2}\footnotemark[1] \quad Zongxia Li\textsuperscript{3}\footnotemark[1] \quad Jiaxin Huang\textsuperscript{2} \quad Yonghui Yang\textsuperscript{4}\\[0.5ex] \textsuperscript{1}University of Illinois Urbana-Champaign \quad \textsuperscript{2}Washington University in St. Louis \\ \textsuperscript{3}University of Maryland \quad \textsuperscript{4}National University of Singapore\\ {\tt\small yh84@uiuc.edu \quad chengsong@wustl.edu \quad zli12321@umd.edu} }

\begin{document}
\maketitle
\begin{abstract}
Reinforcement learning (RL) provides a principled framework for improving Vision-Language Models (VLMs) on complex reasoning tasks. However, existing RL approaches often depend on human-annotated labels or task-specific heuristics to define verifiable rewards—both costly and limited in scalability. 
We introduce \method{}, a self-evolving RL framework that enables VLMs to autonomously improve their reasoning capabilities from massive unlabeled image data. 
Starting from a single base VLM, \method{} assigns the model into two interacting roles: an \textbf{Image-Conditioned Questioner} that formulates challenging yet answerable visual questions, and a \textbf{Multimodal Reasoner} that generates silver responses. 
These roles are jointly trained using Group Relative Policy Optimization (GRPO), which uses diversity and difficulty rewards to balance the difficulty of generated questions with the quality of silver answers. 
\method{} scales efficiently across two model families. 
Trained on Qwen2.5-VL and MiMo-VL, \method{} achieves consistent improvements in visual reasoning, compositional generalization, and hallucination reduction across eight benchmarks including MM-Vet and MMMU, and establishes a scalable path toward self-evolving multimodal intelligence.
Our project page is available at \url{https://bruno686.github.io/VisPlay/}.
% \jh{Just a small suggestion: now the terminology and math equations/symbols seem too alike with the R-zero paper. For example, the role names ``Questioner'' and ``Solver'' looks alike with the ``Challenger'' and ``Solver'' in R-zero. You can consider names like ``Image-Conditioned Questioner'' and ``Multimodal Reasoner'' or other names you think fit.}
\end{abstract}    
\section{Introduction}
Self-evolving mechanisms~\citep{schmidhuber2007godel,clune2019ai} represent a promising frontier for advancing artificial intelligence. The training of state-of-the-art (SoTA) models has traditionally relied on large volumes of expert curated tasks and labels. However, the reliance on human annotation is not only costly, labor-intensive, and difficult to scale, but also presents a fundamental bottleneck to advancing intelligence toward capabilities that could surpass itself without human signal guidance~\citep{yang2024weak}. Self-evolution offers a compelling alternative by equipping models with the capacity to independently generate, refine, and learn from their own experiences such as through self-play or synthetic data generation. 
 
Motivated by these advantages, the research community has increasingly explored self-evolution, most notably in the context of Large Language Models (LLMs).
A line of works have demonstrated how LLMs can autonomously enhance their complex reasoning and coding faculties, often by generating their own tasks or data~\citep{R-zero, liu2025spice, absolute}. 
However, the self-evolution paradigm remains largely underexplored for Vision-Language Models (VLMs)~\cite{qwen2.5-VL, Li_2025_CVPR, bordes2024introductionvisionlanguagemodeling}.
Unlike LLMs, which rely solely on text, developing self-evolving VLMs poses additional challenges due to their dependence on the visual modality.
In a world where human annotation is costly and time-consuming, yet vast amounts of visual data are freely available online, self-evolving VLMs present a promising direction to continual improvement without human signals and directly from the abundant visual content on the internet~\cite{sehgal2025selfevolvingvisualconceptlibrary,chen2025c2evocoevolvingmultimodaldata}.

\begin{figure}[t!] 
  \centering 
  \includegraphics[width=0.45\textwidth]{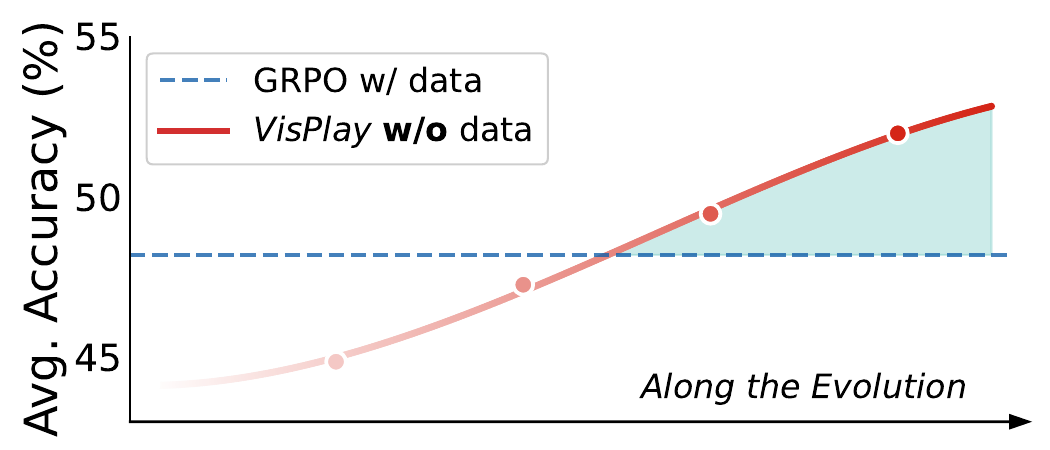}
  \caption{Illustration of the average accuracy improvement (averaged over seven datasets) through successive evolutions (Evo 1 to Evo 5) on Qwen2.5-VL-3B-Instruct, compared to a baseline trained on Vision-47K with GRPO, demonstrating the effectiveness of our \method~.} 
  \label{fig:intro}
\end{figure}

In this paper, we introduce \method, a self-evolving RL framework that enables VLMs to autonomously improve their reasoning capabilities using \textbf{only raw, unannotated images}. The framework utilizes a single base VLM that alternates between two roles: the \textit{Image-Conditioned Questioner}, which generates diverse and challenging questions conditioned on an input image, and the \textit{Multimodal Reasoner}, which produces silver responses based on both the image and the generated question.
Both roles are jointly optimized using Group Relative Policy Optimization (GRPO)~\cite{grpo}, where designed rewards encourage a balance between question difficulty and answer quality without requiring external supervision.
The Image-Conditioned Questioner learns to generate challenging yet answerable questions grounded in visual inputs, while the Multimodal Reasoner learns to produce accurate, detailed, and grounded responses.
This self-evolving framework enables the VLM to progressively improve its visual reasoning abilities through iterative co-improvement of the Questioner and the Reasoner as Figure~\ref{fig:intro}.

We apply our self-evolving RL framework to train three state-of-the-art (SoTA) VLMs and observe consistent performance gains across diverse visual reasoning benchmarks.

Our main contributions are:

\begin{itemize}
\item We propose \method, a self-evolving RL framework for Vision-Language models. 
\item We apply \method~ to three strong models—Qwen2.5-VL-3B, Qwen2.5-VL-7B~\cite{qwen2.5-VL}, and MiMo-VL-7B~\cite{MiMo-VL}. 
We run extensive evaluations over three major domains—General Visual Understanding, Visual Mathematics, and Hallucination Detection. 
All models show consistent gains in accuracy after several iterations. 
\item We run extensive ablation studies to further validate the contribution of Image-Conditioned Questioner and the Multimodal Reasoner component to further show how \method~ progressively strengthens multimodal reasoning across vision-language tasks. 
\end{itemize}

\begin{figure*}[htbp] 
  \centering 
  \includegraphics[width=\textwidth]{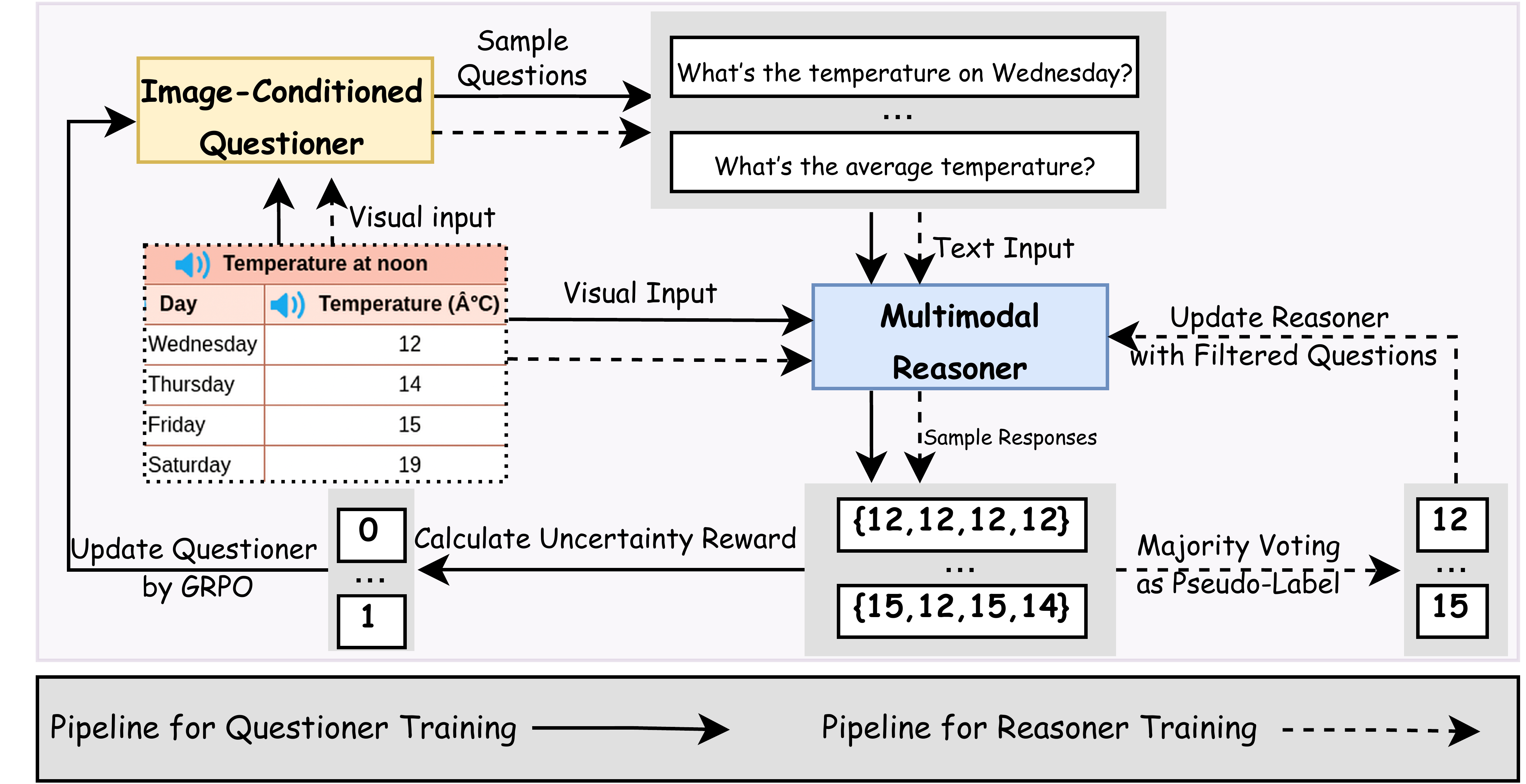}
  \vspace{-0.5cm}
  \caption{An illustration of our \method{} framework, depicting the co-evolution of the Image-Conditioned Questioner and Multimodal Reasoner. Top: During the Questioner training stage, the Image-Conditioned Questioner is optimized via GRPO to produce challenge questions. The reward stems from the uncertainty of the frozen Multimodal Reasoner, computed by the consistency of its multiple generated answers. Bottom: In the Reasoner training stage, the Multimodal Reasoner is trained via GRPO on a curated set of challenging questions from the now-frozen Image-Conditioned Questioner, leveraging pseudo-labels from its own majority voting.} 
  \label{fig:main}
\end{figure*}

\section{Method}
\subsection{Preliminary}
Reinforcement Learning with Verifiable Rewards (RLVR)~\citep{rlvr} 
is a paradigm for training VLMs in domains where the correctness of model outputs can be verified. 
A rule-based verifier $v: X \rightarrow \{0,1\}$ assigns a binary reward to each generation $x_i$:

\begin{equation}
  r_i = v(x_i) =
    \begin{cases}
    1, & \text{if } x_i \text{ satisfies a correctness rule},\\
    0, & \text{otherwise}.
    \end{cases} 
\end{equation}

Such verifiable rewards are effective in tasks like mathematical reasoning, multiple choice, and code generation~\citep{zhao2025one}, 
where correctness can be objectively evaluated. 
GRPO~\citep{grpo} provides a practical RL algorithm without a value function by using relative rewards among multiple samples from the same prompt. Given a prompt $p$, 
a policy $\pi_{\theta_{\text{old}}}$ produces $G$ complete responses 
$\{x_1, \ldots, x_G\}$ with corresponding rewards $\{r_1, \ldots, r_G\}$. 
Rewards are normalized within the group to compute response-level advantages:

\begin{equation}
\label{eq:advantage}
  \hat{A}_i = 
\frac{r_i - \mathrm{mean}(r_1, \ldots, r_G)}
{\mathrm{std}(r_1, \ldots, r_G) + \varepsilon_{\text{norm}}},  
\end{equation}
where $\varepsilon_{\text{norm}}$ is a small constant for stability.

The policy is then optimized using a clipped surrogate objective, 
regularized by a KL term to constrain policy drift:

\begin{equation}
\label{eq:loss_grpo}
 \begin{aligned}
\mathcal{L}_{\text{GRPO}}(\theta)
&= -\frac{1}{G}\!\sum_{i=1}^{G}
\min\!\Bigg(
\frac{\pi_\theta(x_i)}{\pi_{\theta_{\text{old}}}(x_i)} \hat{A}_i,\,
\\[-2pt]
&\quad\mathrm{clip}\!\left(
\frac{\pi_\theta(x_i)}{\pi_{\theta_{\text{old}}}(x_i)},
1\!-\!\epsilon,\, 1\!+\!\epsilon
\right)
\hat{A}_i
\Bigg)
\\[2pt]
&\quad+ \beta\, \mathrm{KL}\!\left(\pi_\theta \,\|\, \pi_{\theta_{\text{old}}}\right).
\end{aligned}   
\end{equation}
GRPO operationalizes RLVR principles to improve reasoning and generation quality in VLMs by rewarding responses with positive relative advantages while limiting policy deviation

\subsection{Pipeline Overview}
We introduce \method, a self-play reinforcement learning framework designed to evolve VLMs without human-annotated data. As illustrated in Figure~\ref{fig:main}, the framework operates as a closed-loop system involving two agents evolved from the same base model: an \emph{Image-Conditioned Questioner} and a \emph{Multimodal Reasoner}.
The process begins with the Questioner taking an image as input to generate a visual query. Subsequently, the Reasoner receives both the image and the generated query to produce a response. Both the Questioner and the Reasoner are initialized from a shared pretrained backbone. The two agents co-evolve through iterative interactions: the Questioner is trained to generate more challenging questions, while the reasoner is trained to solve more and more challenging questions. The complete process is described in Algorithm~\ref{alg:self_play}.

\subsection{Image-Conditioned Questioner Training}
\label{sec:questioner_training}
The Questioner is an autoregressive policy denoted by \(Q_{\theta}\). Conditioned on an input image \(I\), it samples a group of \(G\) questions \(\{x_i\}_{i=1}^G \sim Q_{\theta}(\cdot|I)\), which are evaluated to produce scalar rewards \(\{r_i\}_{i=1}^G\). These rewards are used to compute group-normalized advantages and to update \(Q_{\theta}\) with a GRPO objective.
We next define reward components that constitute each \(r_i\).

\paragraph{Pseudo-Label Generation.}
Since self-evolving VLMs learn without relying on labeled data, ground-truth answers for the Questioner’s generated questions are unavailable. Therefore, we introduce a method to approximate the corresponding answers.
Given an image \(I\) and the generated question \(x\), we introduce a Reasoner \(S_{\phi}\) that samples \(m\) responses \(\{y_j\}_{j=1}^m\).\footnote{The reasoner is evolved from the same base model as the Questioner. See Section~\ref{sec:reasoner_training} for details.}
We define the empirical frequency of a candidate answer \(y\) as 
\(\hat{p}(y | x, I) = \frac{1}{m}\sum_{j=1}^m \mathbbm{1}\{y_j = y\}\),
and derive the pseudo-label via majority voting~\citep{huang2025efficient}: 
\(\tilde{y} = \arg\max_{y}\hat{p}(y | x, I)\).
We then define the \textbf{confidence score} for this pseudo-label as:
\begin{equation}
\label{eq:conf}
\text{conf}(x, I) = \hat{p}(\tilde{y} | x, I).
\end{equation}
Intuitively, \(\text{conf}(x, I)\) measures the Reasoner's certainty about the pseudo-label: 
high values indicate stable and consistent predictions, whereas values near 0.5 reflect strong uncertainty. 
We therefore treat the degree of uncertainty (i.e., how close \(\text{conf}(x, I)\) is to 0.5) as a proxy for the model-perceived difficulty of the generated question.

\paragraph{Uncertainty Reward.}
The confidence score quantifies the Reasoner’s uncertainty, which we use as a proxy for the model-perceived difficulty of the generated question.
To encourage questions that probe the Reasoner's limits, we compute the reward based on the confidence score \(c = \text{conf}(x, I)\). We define the uncertainty reward to penalize deviations from the point of maximum uncertainty:
\begin{equation}
r_{\mathrm{unc}}(x, I) \;=\; 1 - \bigl| 2c - 1 \bigr|.
\end{equation}
This formulation yields a maximal reward of 1 when \(c = 0.5\) and decreases linearly to 0 as the reasoner's response distribution becomes deterministic (i.e., \(c \to 1\)).

\paragraph{Diversity Regularization.}
To prevent the model from collapsing~\citep{dai2025cde,chen2025pass,zhou2025evolving,liu2025vogue,cheng2025reasoning} into generating repetitive questions for a given image \(I\), we introduce a redundancy penalty within its generated group \(\mathcal{X}_I\).
We cluster these generated questions based on pairwise similarity (BLEU score) to identify duplicates. For a question \(x_i\) belonging to a cluster \(C_k^{(I)} \subseteq \mathcal{X}_I\), the regularization term is:
\begin{equation}
\label{cluster}
r_{\mathrm{div}}(x_i, I) \;=\; \lambda \, \frac{|C_k^{(I)}|}{G},
\end{equation}
where \(C_k^{(I)}\) denotes the cluster of similar questions for image \(I\), and \(G\) is the total number of generated questions for that image.

\paragraph{Format Constraint.}
We enforce a hard filter to ensure structural validity. Specifically, we require the generated question to be strictly enclosed within \texttt{<question>} tags. Any output failing to meet this format requirement is assigned zero reward. We denote this validity indicator as:
\begin{equation}
\mathbbm{1}_{\mathrm{valid}}(x) =
\begin{cases}
1, & \text{if } x \text{ is wrapped in } \texttt{<question>} \text{ tags},\\
0, & \text{otherwise}.
\end{cases}
\end{equation}

\paragraph{Final Questioner Reward.}
For each generated question \(x_i\) conditioned on image \(I\), we integrate the uncertainty signal and diversity regularization into a unified scalar objective:
\begin{equation}
r_i \;=\; \mathbbm{1}_{\mathrm{valid}}(x_i) \cdot \mathrm{ReLU}\Bigl(r_{\mathrm{unc}}(x_i, I) - r_{\mathrm{div}}(x_i, I)\Bigr).
\end{equation}
This composite reward incentivizes the Questioner to generate challenging yet non-redundant questions while strictly filtering out malformed outputs. The ReLU function stabilizes GRPO updates by preventing spurious negative values from skewing the reward normalization across the group.

\subsection{Multimodal Reasoner Training}
\label{sec:reasoner_training}
The training of the Multimodal Reasoner \(S_{\phi}\) builds upon the advancements of the Image-Conditioned Questioner. In each iteration, the Image-Conditioned Questioner functions produce challenging samples that serve as training targets. The Multimodal Reasoner then learns from these automatically curated samples, improving its visual reasoning ability without any external supervision.

\begin{algorithm}[t]
\caption{\method: Self-Evolving RL for Vision-Language Models}
\label{alg:self_play}
\SetAlgoLined
\KwInput{Initial models $Q_\theta, S_\phi$; Image dataset $\mathcal{I}$; Group size $G$; Reasoner samples $m$; Dataset budget $N$; Thresholds $\tau_{\mathrm{low}}=0.25, \tau_{\mathrm{high}}=0.75$.}
\KwOutput{Evolved models $Q_\theta$ and $S_\phi$.}

\For{each self-play iteration}{
    \For{each image batch $I \in \mathcal{I}$}{
        Sample question group $\{x_i\}_{i=1}^G \sim Q_\theta(\cdot \mid I)$\;
        \For{each question $x_i$}{
            Sample $m$ answers $\{y_j\}_{j=1}^m \sim S_{\phi}(\cdot \mid I, x_i)$\;
            Compute confidence $c_i \leftarrow \text{conf}(x_i, I)$ via majority vote (Eq.~\ref{eq:conf})\;
            Compute uncertainty reward $r_{\mathrm{unc}} \leftarrow 1 - |2c_i - 1|$\;
            Compute diversity penalty $r_{\mathrm{div}}$ via clustering (Eq.~\ref{cluster})\;
            Final reward: $r_i \leftarrow \mathbbm{1}_{\mathrm{valid}}(x_i) \cdot \mathrm{ReLU}\bigl(r_{\mathrm{unc}} - r_{\mathrm{div}}\bigr)$\;
        }
        Update $Q_\theta$ via GRPO using rewards $\{r_i\}_{i=1}^G$\;
    }
    Initialize curated dataset $\mathcal{S} \leftarrow \emptyset$\;
    \For{each image $I \in \mathcal{I}$}{
        Generate $N$ candidate questions via $Q_\theta$ \;
        \For{each candidate $x_k$}{
             Obtain pseudo-label $\tilde{y}_k$ and confidence $c_k$ from $S_\phi$\;
             \If{$\tau_{\mathrm{low}} \le c_k \le \tau_{\mathrm{high}}$}{
                Add $(I, x_k, \tilde{y}_k)$ to $\mathcal{S}$\;
                }
        }
    }
    
    \For{each minibatch $(I, x, \tilde{y}) \in \mathcal{S}$}{
        Sample $G$ answers $\{y_j\}_{j=1}^G \sim S_\phi(\cdot \mid I, x)$\;
        Compute binary rewards $r_j \leftarrow \mathbbm{1}(y_j = \tilde{y})$\;
        Update $S_\phi$ via GRPO using rewards $\{r_j\}_{j=1}^G$\;
    }
}
\end{algorithm}

\paragraph{Curated Dataset Construction.}
\label{sec:data_gen}
Following the update of the Image-Conditioned Questioner, we generate a diverse pool of \(N\) candidate questions \(\{x_i\}_{i=1}^N\) per image by sampling \(x_i \sim Q_{\theta}(\cdot\mid I)\). For each \(x_i\), we obtain \(m\) response samples from the current Multimodal Reasoner and compute the pseudo-label \(\tilde{y}_i\) and confidence score \(c_i = \text{conf}(x_i, I)\). 
To focus on training samples that offer high information gain, we enforce an \emph{informative filter} that retains pairs \((x_i, \tilde{y}_i)\) with moderate confidence:
\begin{equation}
\tau_{\mathrm{low}} \le c_i \le \tau_{\mathrm{high}},
\end{equation}
where \(\tau_{\mathrm{low}}\) and \(\tau_{\mathrm{high}}\) are thresholds set to 0.25 and 0.75, respectively. This criterion effectively discards trivial samples where the model is already certain (\(c_i > 0.75\)) as well as highly unstable or noisy generations (\(c_i < 0.25\)). The final curated training set \(\mathcal{S}\) is formed by collecting all retained pairs across images, up to a budgeted size, to optimize the Multimodal Reasoner via GRPO.

\begin{table*}[t!]
\centering
\caption{Comprehensive results on visual reasoning benchmarks. Each base model is evaluated against two settings: a \method{} (\raisebox{-0.3ex}{\includegraphics[width=0.02\textwidth]{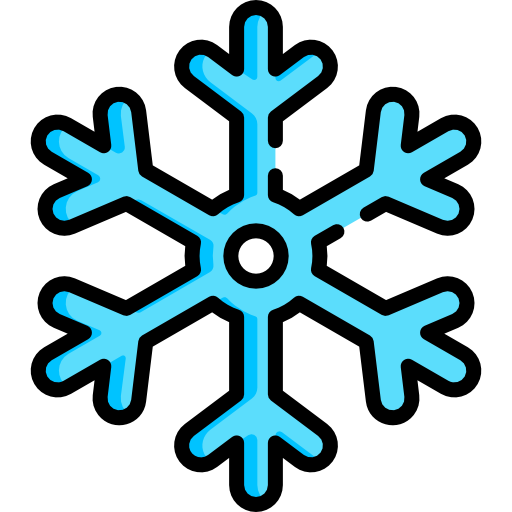}} challenger) baseline, in which the Reasoner is trained on questions produced by an untrained Challenger, and our iterative \method{} framework. The highest performance reached during training for each model is emphasized in bold. We take accuracy as the metric.}
\vspace{-0.1in}
\label{tab:final_main_results}
\resizebox{0.9\textwidth}{!}{
\begin{tabular}{@{}lccccccccc@{}}
\toprule
    & \multicolumn{4}{c}{\textbf{General Visual Understanding}} & \multicolumn{2}{c}{\textbf{Visual Math}} & \textbf{Hallucination} & \\
    \cmidrule(lr){2-5}\cmidrule(lr){6-7}\cmidrule(lr){8-8}
    \multirow{2}{*}{\textbf{Methods}} &  \multirow{2}{*}{\textbf{MMMU}} & \textbf{MM} & \textbf{RealWorld} & \textbf{VisNum} & {\textbf{Math}} & \textbf{MATH} & \textbf{Hallusion} & \multirow{2}{*}{\textbf{Avg.}} \\
    & & \textbf{-Vet} & \textbf{QA} & \textbf{Bench} & \textbf{Verse} & \textbf{-Vision} & \textbf{Bench} & \\
\midrule
\multicolumn{9}{@{}l}{\textit{Qwen2.5-VL-3B-Instruct}} \\
\quad Base Model     & 19.95 & 36.24 & 49.28 & 27.08 & 26.14 & 20.23 & 32.81 & 30.61 \\
\quad \method{} (\raisebox{-0.3ex}{\includegraphics[width=0.02\textwidth]{figures/snowflake.png}} challenger)  & 23.34 & 43.58 & 57.78 & 29.33 & 33.50 & 23.39 & 64.88 & 33.77\\
\quad \method{} (Iter 1)& 29.40 & \textbf{48.62} & 67.06 & 30.01 & 29.67 & 22.57 & 91.80 & 44.16 \\ 
\quad \method{} (Iter 2)&  33.37 & 44.50 & 65.62 & 29.64 & 32.36 & 24.67 & \textbf{94.95} & 44.87 \\
\quad \method{} (Iter 3)&   \textbf{37.11} & 38.07 & \textbf{71.90} & \textbf{39.15} & \textbf{35.15} & \textbf{29.97} & 90.54 & \textbf{47.27} \\
\midrule
\multicolumn{9}{@{}l}{\textit{Qwen2.5-VL-7B-Instruct}} \\
\quad Base Model     &   23.10 & 44.95 & 57.52 & 32.57 & 33.78 & 24.05 & 66.88 & 40.41 \\
\quad \method{} (\raisebox{-0.3ex}{\includegraphics[width=0.02\textwidth]{figures/snowflake.png}} challenger)  &35.24 & 45.87 & 69.67 & 32.41 & 35.13 & 26.22 & 78.13 & 38.33 \\
\quad \method{} (Iter 1)&  28.94 & 46.33 & 62.61 & 28.65 & 33.88 & 26.91 & 80.34 & 44.53 \\
\quad \method{} (Iter 2)&   27.07 & 42.66 & 60.92 & 27.08 & 36.32 & 25.00 & 67.72 & 40.97 \\
\quad \method{} (Iter 3)&  \textbf{38.27} & \textbf{46.33} & \textbf{69.67} & \textbf{32.57} & \textbf{39.14} & \textbf{31.15} & \textbf{92.32} & \textbf{48.61} \\
\midrule
\multicolumn{9}{@{}l}{\textit{MiMo-VL-7B-SFT}} \\
\quad Base Model     &  30.22 & \textbf{59.17} & 78.17 & 44.80 & 41.80 & 25.33 & 87.17 & 43.56 \\
\quad \method{} (\raisebox{-0.3ex}{\includegraphics[width=0.02\textwidth]{figures/snowflake.png}} challenger)  &27.54 & 58.72 & 63.27 & 49.66 & 38.78 & 24.80 & 57.83 & 39.63 \\
\quad \method{} (Iter 1)&  25.67 & 56.42 & 69.54 & 51.59 & 40.20 & 25.13 & \textbf{87.59} & 43.16 \\
\quad \method{} (Iter 2)&  \textbf{34.07} & 55.96 & \textbf{78.69} & 51.18 & 41.65 & 28.45 & 86.65 & 45.58 \\
\quad \method{} (Iter 3)& 28.24 & 56.88 & 71.50 & \textbf{52.69} & \textbf{46.02} & \textbf{29.44} & 74.55 & \textbf{45.69}\\
\bottomrule
\end{tabular}}
\end{table*}

% \quad \method{} (Iter 4)&   45.04 & 34.86 & 74.38 & 63.51 & 41.09 & 31.91 & 99.47 & 49.48 \\
% \quad \method{} (Iter 5)&   40.96 & 32.11 & 68.63 & 54.94 & 37.31 & 32.96 & 97.06 & 51.99 \\

\paragraph{Per-Sample Verifiable Reward.}
For a question \(x_i\in\mathcal{S}\) with pseudo-label \(\tilde{y}_i\), the Multimodal Reasoner generates a group of \(G\) candidate answers \(\{y_j\}_{j=1}^G\). Each sampled answer receives the binary reward
\begin{equation}
r_j =
\begin{cases}
1, & \text{if } y_j = \tilde{y}_i,\\[2pt]
0, & \text{otherwise}.
\end{cases}
\end{equation}
These rewards are group-normalized to produce advantages \(\hat{A}_j\) as in Eq.~\ref{eq:advantage} (with the Reasoner’s rewards), and \(S_{\phi}\) is updated by minimizing \(\mathcal{L}_{\mathrm{GRPO}}(\phi)\) as in Eq.~\ref{eq:loss_grpo}.
\section{Experiments}

\subsection{Benchmarks and Evaluation Protocol}
We use existing image datasets from Vision-47K~\cite{vision-SR, feng2025videor1reinforcingvideoreasoning}, which contains 47K web images collected from diverse domains, e.g. including charts, medical images, exams, textbooks, and driving simulations.\footnote{We only use the images without the questions and answers. Details of the dataset breakdowns are in supplement materials.}
We train three backbone models using \method{}—Qwen2.5-VL-3B-Instruct, Qwen2.5-VL-7B-Instruct, and Mimo-7B-SFT.\footnote{The detailed training configurations are provided in the supplementary material.}
We run evaluation across three multimodal domains~\cite{vision-SR}.
\begin{itemize}
    \item \textbf{General Visual Understanding.} We measure performance on four established benchmarks. MM-Vet~\cite{Mm-vet} provides a unified LLM-based score across recognition, OCR, and visual math tasks. 
    MMMU~\cite{mmmu} evaluates cross-modal reasoning and subject knowledge through 11.5K college-level, four-choice questions spanning six academic disciplines.
    RealWorldQA~\cite{RealWorldQA2024} contains roughly 700 real-world images paired with spatially grounded questions. VisNumBench~\cite{VisNumBench} focuses on visual number sense, covering around 1.9K questions involving numerical attributes and estimation tasks.
    \item \textbf{Multimodal Mathematical Reasoning.} MathVerse~\cite{zhang2024mathversedoesmultimodalllm} consists of 2.6K diagram-centric questions spanning geometry and functions, provided in multiple visual-text formats. MATH-Vision~\cite{wang2024measuringmultimodalmathematicalreasoning} includes around 3K competition-level problems across 16 subjects and five difficulty tiers.
    \item \textbf{Visual Hallucination Detection.} HallusionBench~\cite{Guan_2024_CVPR} is used to analyze model errors, distinguishing between language-only hallucinations and visual-illusion errors, with a simple yes/no evaluation format.
\end{itemize}

\begin{figure*}[h]
    \centering
    \begin{subfigure}[b]{0.31\textwidth}
        \centering
        \includegraphics[width=\textwidth]{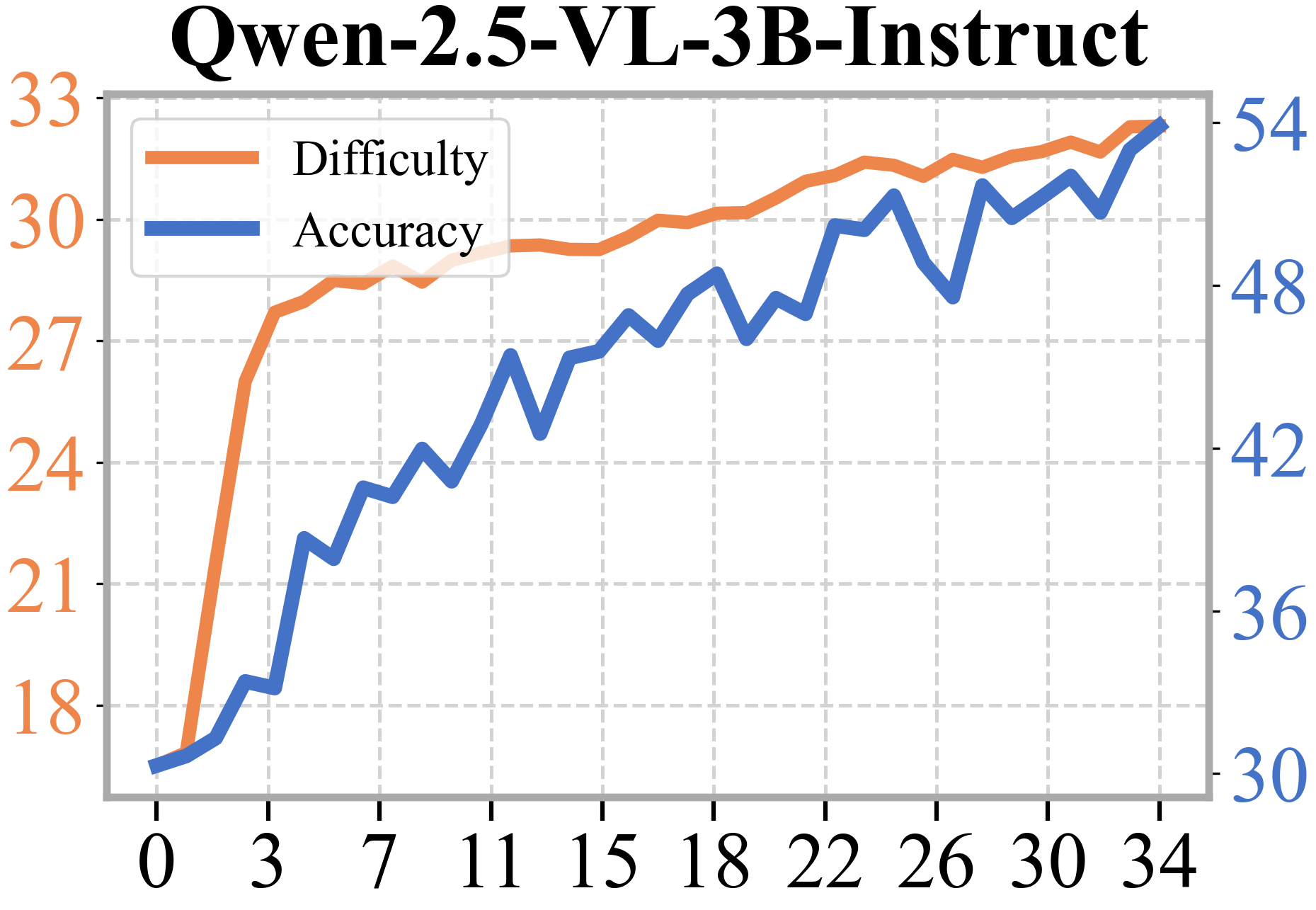}
        \label{fig:sub4}
    \end{subfigure}
    \hfill
    \begin{subfigure}[b]{0.34\textwidth}
        \centering
        \includegraphics[width=\textwidth]{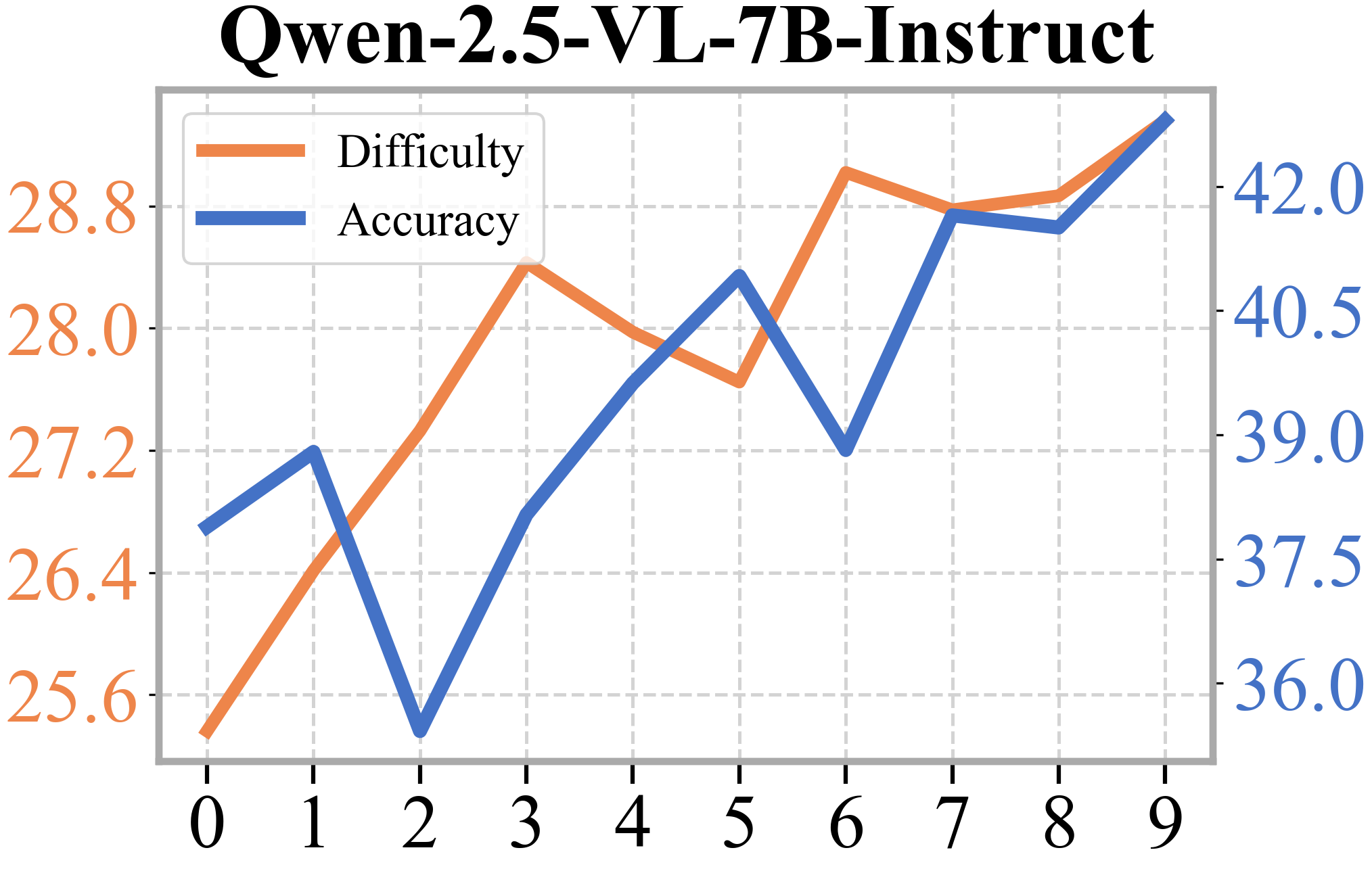}
        \label{fig:sub5}
    \end{subfigure}
    \hfill
    \begin{subfigure}[b]{0.33\textwidth}
        \centering
        \includegraphics[width=\textwidth]{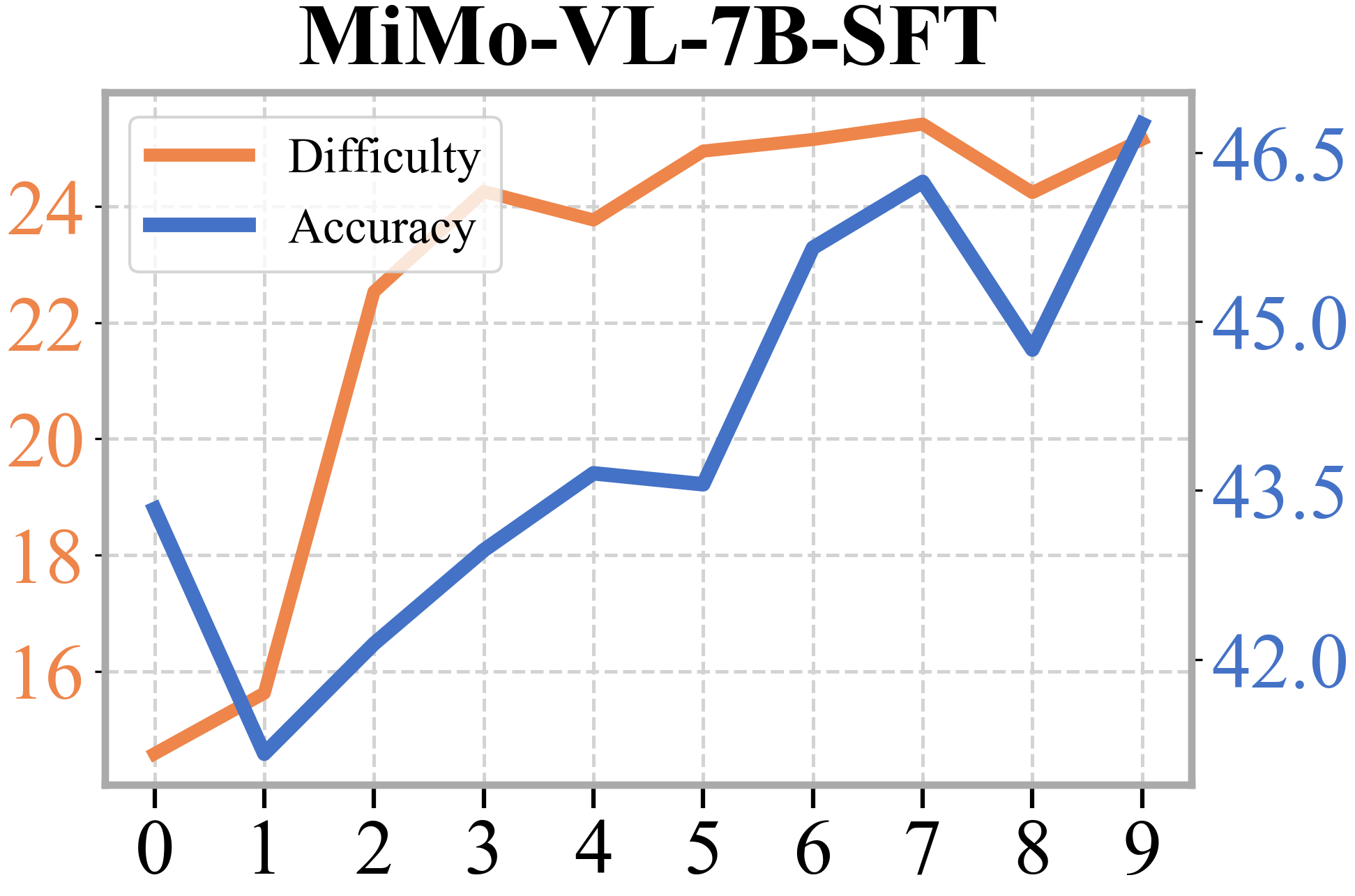}
        \label{fig:sub6}
    \end{subfigure}
    \vspace{-0.1in}
    \caption{Changes in question difficulty (orange, left axis) and problem-solving accuracy (blue, right axis) 
    during Image-Conditioned Questioner and Multimodal Reasoner training across three VLMs.}
    \label{fig:diffcult&acc}
\end{figure*}

\begin{table*}[h!]
    \centering
    \caption{Analysis of model performance and data quality. The shaded column indicates the estimated accuracy of the self-generated pseudo-labels for each question set, as determined using ChatGLM-Flash.}
    \label{tab:evolution}
    \vspace{-0.1in}
    \begin{tabular}{@{}lcccc>{\columncolor{gray!15}}c@{}}
        \toprule
        & \multicolumn{5}{c}{\textbf{Performance of Evaluated Model (vs. Ground Truth)}} \\
        \cmidrule(l){2-6}
        & Base Model & Reasoner (Iter 1) & Reasoner (Iter 2) & Reasoner (Iter 3) & Pseudo-Label Acc. \\
        \midrule
        $\mathcal{D}_{\text{Iter 1}}$ & 39.0 & 44.0 & 45.5 & 49.0 & 72.0 \\
        $\mathcal{D}_{\text{Iter 2}}$ & 37.5 & 42.5 & 44.0 & 47.5 & 65.0 \\
        $\mathcal{D}_{\text{Iter 3}}$ & 36.0 & 40.0 & 41.5 & 45.0 & 61.0 \\
        \bottomrule
    \end{tabular}
\end{table*}

\subsection{Main Results}
We use LLM-as-a-judge to assess the correctness of the answers to ensure more robust evaluation~\cite{gu2025surveyllmasajudge, li-etal-2025-llm-judge}. 
We present the outputs of the Multimodal Reasoner and analyze its reasoning ability progression in Table~\ref{tab:final_main_results}.
We summarize the main findings of our experimental results below.
% The experimental results \method{}, on visual reasoning benchmarks are summarized in Table~\ref{tab:final_main_results}. Furthermore, we summarize the main findings as follows.

\begin{itemize}
\item \textbf{\method{} consistently improves overall performance across different models.}
All models trained with \method{} consistently surpass both their corresponding base models and the Base Challenger over successive training iterations. Qwen2.5-VL-3B shows a remarkable improvement, with the average score increasing from 30.61 at baseline to 44.16 after the first iteration and reaching 47.27 at the third. Qwen2.5-VL-7B and MiMo-VL-7B follow similar upward trends, improving from 40.41 to 48.61 and 43.56 to 45.69, respectively.
These results demonstrate the robust generalization ability and scalability of the proposed self-evolving framework across different models and model sizes.

\item \textbf{Performance gains across diverse task types.}
\method{} shows improvements across general visual understanding tasks, visual reasoning or math benchmarks, and are more robust to hallucination. For Qwen2.5-VL-3B, the Hallucination score rises from 32.81 to 94.95 by the second iteration, showing a substantial enhancement in factual grounding. Similar patterns are observed in other models—reasoning benchmarks consistently improve without compromising accuracy on general understanding tasks—demonstrating that \method{} effectively strengthens both task-specific reasoning and cross-domain multimodal generalization.

\item \textbf{Iterative co-evolution between the Questioner and Reasoner drives improvement.}
Performance trajectories across iterations highlight the co-evolution between the Questioner and Reasoner. As the Questioner generates more diverse and challenging queries, the Reasoner—trained with GRPO using high-quality silver supervision—learns to handle increasingly complex reasoning steps. This iterative loop allows both components to reinforce each other, leading to continual improvement in reasoning quality, generalization, and robustness. The results indicate that the co-evolutionary design of \method{} provides a scalable path toward self-improving multimodal intelligence.
\end{itemize}

\begin{table*}[th!]
\centering
\caption{Performance comparison between \method{} and standard GRPO training with human-labeled data. Although \method{} relies entirely on self-generated supervision, it achieves competitive overall accuracy and significantly reduces hallucination. This demonstrates that our self-evolving framework can meaningfully enhance VLM performance even in the absence of human-annotated datasets.}
\vspace{-0.1in}
\label{tab:abalation_results}
\resizebox{0.9\textwidth}{!}{
\begin{tabular}{@{}lccccccccc@{}}
\toprule
    & \multicolumn{4}{c}{\textbf{General Visual Understanding}} & \multicolumn{2}{c}{\textbf{Visual Math}} & \textbf{Hallucination} & \\
    \cmidrule(lr){2-5}\cmidrule(lr){6-7}\cmidrule(lr){8-8}
    \multirow{2}{*}{\textbf{Methods}} &  \multirow{2}{*}{\textbf{MMMU}} & \textbf{MM} & \textbf{RealWorld} & \textbf{VisNum} & {\textbf{Math}} & \textbf{MATH} & \textbf{Hallusion} & \multirow{2}{*}{\textbf{Avg.}} \\
    & & \textbf{-Vet} & \textbf{QA} & \textbf{Bench} & \textbf{Verse} & \textbf{-Vision} & \textbf{Bench} & \\
\midrule
\multicolumn{9}{@{}l}{\textit{Qwen2.5-VL-3B-Instruct}} \\
\quad Standard GRPO & 40.3 & 49.5 & 63.0 & 36.7 & 42.8 & 29.9 & 67.4 & 47.1 \\
\quad \method{} (Iter 3)& 37.1 & 38.1 & 71.9 & 39.2 & 35.2 & 30.0 & 90.5 & 47.3 \\
\midrule
\multicolumn{9}{@{}l}{\textit{Qwen2.5-VL-7B-Instruct}} \\
\quad Standard GRPO & 39.8 & 51.8 & 66.6 & 43 & 53.2 & 33.8 & 66.6 & 50.7 \\
\quad \method{} (Iter 3)& 38.3 & 46.3 & 69.7 & 32.6 & 39.1 & 31.2 & 92.3 & 48.6 \\
% \midrule
\bottomrule
\end{tabular}}
\end{table*}

\begin{table*}[h!]
\centering
\caption{Examples of challenging questions generated by the self-evolving Vision-Language model across three training iterations. The questions progressively increase in complexity, illustrating the growth in difficulty of the Questioner’s outputs over iterations. Images on the left and right correspond to the visual context for each question. The questions are raised by Qwen2.5-VL-3B-Instruct.}
\begin{adjustbox}{width=\textwidth}
\renewcommand{\arraystretch}{1.5} % 增加行高
\begin{tabularx}{\textwidth}{
    >{\centering\arraybackslash}m{0.15\textwidth}  % 左列：Iteration 居中
    >{\noindent\justifying\arraybackslash}X  % 中列：文字两端对齐
    >{\noindent\justifying\arraybackslash}X  % 右列：文字两端对齐
}
\toprule
\multicolumn{3}{l}{\textbf{Challenging Examples from Self-Evolving Trained Vision-Language Model}} \\
\midrule
 & \includegraphics[width=0.8\linewidth]{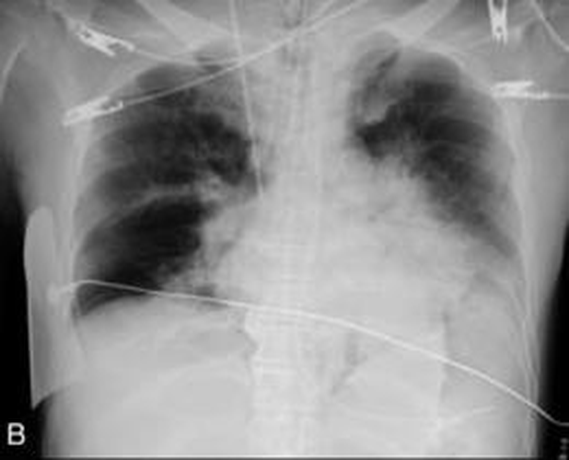} 
& \includegraphics[width=0.8\linewidth]{figures/f2.png} \\
\midrule
Question (Iter 1) & 
Approximately how many \textcolor{red!70!black}{lung fields} are visible in the X-ray image? 
& Which \textcolor{red!70!black}{skeletal structure} most likely belongs to a bird with \textcolor{red!70!black}{hollow bones}? \\
\midrule
Question (Iter 2) & 
On a thoracic x-ray, the right lobe of the lung is more spread out compared to the left lobe. If the right lobe is given a score of 1 and the left lobe is given a score of 0, what is the \textcolor{red!70!black}{difference in scores} between the right and left lung lobes?
& On which figure does the \textcolor{red!70!black}{long neck of the dinosaur} have the greatest \textcolor{red!70!black}{horizontal angle} with the vertical axis? \\
\midrule
Question (Iter 3) & 
On which \textcolor{red!70!black}{rib} is the line approximately \textcolor{red!70!black}{2.5 cm above the image's midpoint}? 
& Which \textcolor{red!70!black}{skeletal structure} is most likely to have evolved \textcolor{red!70!black}{secondary to flying abilities} and which is less likely to have this trait? \\
\bottomrule
\end{tabularx}
\label{case}
\end{adjustbox}
\end{table*}

\subsection{Performance Comparison with Human-Annotated Data}
We conduct a performance comparison between models trained with \method{} and those trained using human-curated image–question–answer pairs from the Vision-47K dataset under standard GRPO for one epoch, as shown in Table~\ref{tab:abalation_results} for Qwen2.5-VL-3B and 7B. Although this experiment is not an ablation study in the strict sense, it provides a clear view of how our fully automated training pipeline performs relative to conventional supervised training.
Overall, we observe that models trained with \method{} achieve competitive average accuracy compared with those trained on real, human-written data. While performance on several task categories differs slightly, the general trend indicates that the self-evolving process can produce training signals of sufficient quality to improve base VLMs capabilities. These findings suggest that even in settings where human annotations are costly, limited, or unavailable, our framework can still serve as an effective and scalable alternative, enabling VLMs to develop stronger generalization abilities without depending on manual supervision.

\subsection{Co-Evolution Dynamics of Two Roles}
\begin{itemize}
    \item \textbf{The Evolution of Question Difficulty and Solution Accuracy.} To analyze the co-evolution dynamics of the two roles, we examine the changes in question difficulty (orange, left axis) and problem-solving accuracy (blue, right axis) across three VLMs during the first training iteration (Figure~\ref{fig:diffcult&acc}). Question difficulty is operationalized as the Reasoner's model-perceived difficulty, derived from the confidence score defined in Eq.~\ref{eq:conf} in Section~\ref{sec:questioner_training}. Across all models, a consistent co-evolution pattern emerges. The Image-Conditioned Questioner’s difficulty curves exhibit a general upward trend, with initial increments followed by sustained growth, indicating the role’s ability to progressively formulate more challenging visual questions. Concurrently, the Multimodal Reasoner’s accuracy curves, despite minor fluctuations, show a complementary upward trajectory. This means that as question difficulty rises, the Reasoner adapts and enhances its problem-solving capability, with both metrics reinforcing each other’s improvement over iterations. Such mutual reinforcement validates \method{}’s core mechanism, where the two interacting roles drive scalable self-evolution in multimodal reasoning.
    \item \textbf{The Evolution of Capabilities and Data Accuracy.}
    Building on the interaction between question difficulty and solution accuracy, we further analyze how the Reasoner’s capabilities and the quality of pseudo-labeled data evolve across iterations. For each training iteration, the Questioner generates questions for the same 200 images, and Reasoners from each iteration attempt to answer them. As shown in Table~\ref{tab:evolution}, the Reasoner’s accuracy steadily improves across iterations (e.g., from 44.0 to 49.0 on first-iteration questions), while the estimated accuracy of pseudo-labels slightly declines (from 72 to 61), reflecting increasing question difficulty. These trends highlight the co-evolution of model reasoning ability and data complexity during self-improving training.
\end{itemize}

\subsection{Case Study on Question Difficulty Evolution}
Table~\ref{case} presents example questions generated by the self-evolving Vision-Language Models across three training iterations. Iteration 1 questions focus on direct observation, such as counting or identifying objects. Iteration 2 introduces relational and comparative reasoning, requiring the model to assess differences or evaluate spatial angles. Iteration 3 further increases complexity with multi-step reasoning and inference, including precise localization and causal relationships. This progression demonstrates a systematic increase in question difficulty, providing increasingly challenging training signals that encourage the model to adapt and improve its reasoning capabilities. Such a design ensures that both the Questioner and Reasoner co-evolve, progressively enhancing the overall performance of the system.

\section{Related Work}
\paragraph{Post‑Training for Vision‑Language Models}
Recent research in post-training of vision-language models (VLMs) has shifted from supervised fine-tuning (SFT) toward reinforcement learning (RL) paradigms, driven by the increasingly strong capabilities of pre-trained VLMs.
Earlier works such as LLaVA~\cite{liu2023visualinstructiontuning,jia2024describethenreasonimprovingmultimodalmathematical} primarily rely on SFT to align a language model backbone with a visual information via a projection layer, enabling multimodal instruction-following and visual reasoning.
However, as base model quality improves, RL-based post-training has emerged as a more powerful alternative. In particular, R1-style training~\cite{deepseekai2025deepseekr1incentivizingreasoningcapability,zheng2025parallel,jin2025search} has gained attention for its ability to enhance reasoning and visual understanding without explicit supervision.
A key insight behind this success is that RL becomes effective only when the base model is sufficiently capable to self-explore reasoning trajectories~\cite{li2022blipbootstrappinglanguageimagepretraining, Li_2025_CVPR, bai2023qwenvlversatilevisionlanguagemodel, chu2024visionllamaunifiedllamabackbone}.
Despite these advances, most existing RL-based VLM approaches still depend on annotated multimodal datasets, which are costly and difficult to scale~\cite{feng2025videor1reinforcingvideoreasoning, xia2025visionaryr1mitigatingshortcutsvisual, fan2025sophiavlr1reinforcingmllmsreasoning, li2025videohalluevaluatingmitigatingmultimodal,  zhang2025r1vllearningreasonmultimodal, peng2025lmmr1empowering3blmms}.
To reduce reliance on human supervision, several recent works explore VLM self-play paradigms in games. Vision-Zero~\cite{wang2025visionzeroscalablevlmselfimprovement} and Game-RL~\cite{tong2025gamerlsynthesizingmultimodalverifiable} train VLMs with simulated game data to improve their general reasoning ability.
Nonetheless, these methods often continue to depend on external models or tools for training data generation.
While large language models (LLMs) have demonstrated self-evolving learning dynamics without any human or model-labeled data, extending such paradigms to VLMs remains more challenging due to the additional visual modality.
%

% pre‑training–fine‑tuning paradigm toward a dedicated post‑training phase aimed at bridging gaps in modality alignment, reasoning depth and interactive capabilities. Traditional vision‑language pre‑training excels at static tasks such as image captioning or retrieval, but often falls short when deployed on complex downstream multimodal reasoning or interactive usage. To address this, workflows perform post‑training by introducing large image–text (or video–text) corpora, vision‑centric self‑supervised tasks (for example, image patch ordering or region retrieval) and further cross‑modal generation objectives, thereby strengthening both visual structure understanding and textual generation. Despite the gains in robustness, domain adaptability and multimodal expressivity, several open issues persist: designing post‑training tasks that enhance long‑horizon multimodal reasoning without sacrificing generality, adapting effectively when downstream domain data are scarce, and avoiding catastrophic forgetting of the model’s broad multimodal capacity during specialisation.

\paragraph{Self-Evolving In Large Language Models}

Recent work has focused on enabling large language models (LLMs) to self-evolve their reasoning capabilities with minimal to zero human supervision. Various approaches have been proposed to achieve this. General unsupervised self-training frameworks like Genius~\citep{xu2025genius} and Deep Self-Evolving Reasoning~\citep{liu2025deep} aim for advanced reasoning. A common theme is data-free training or starting from zero data, as explored in R-Zero~\citep{R-zero}, Absolute zero~\citep{absolute}, and Language self-play~\citep{kuba2025language}. Other methods adapt self-play for specific goals or environments. For example, SPICE~\citep{liu2025spice} improves reasoning in corpus environments, Search Self-play~\citep{lu2025search} pushes capabilities via search, and SPELL~\citep{yang2025spell} focuses on evolving long-context models. Additionally, some methods utilize interactions between multiple agents to collectively bootstrap reasoning abilities, as seen in Socratic-Zero~\citep{wang2025socratic} and Multi-Agent Evolve~\citep{chen2025multi}.
\section{Limitation}
While our work introduces a scalable, self-evolving framework, we acknowledge two primary limitations that suggest directions for future research.
First, due to computational constraints, our experiments were limited to the Qwen2.5-VL and MiMo-VL families. The scalability and effectiveness of \method{} on significantly larger VLMs (e.g., $\geq$ 10B parameters) is still an important open question.
Second, our framework lacks a definitive verification method for the self-generated data. While our GRPO policy indirectly optimizes for quality, developing more robust, automated methods to verify data faithfulness and prevent error accumulation is a key area for future investigation.

\section{Conclusion}
We present \method, a self-evolving RL framework that enables vision-language models to autonomously improve from unlabeled images. By decomposing a VLM into an Image-Conditioned Questioner and a Multimodal Reasoner and optimizing them via GRPO, our method balances challenge and accuracy without human supervision. Experiments show consistent gains in reasoning, compositional generalization, and hallucination reduction across multiple benchmarks.
\method{} demonstrates that scalable, self-improving multimodal intelligence is achievable. By iteratively generating and learning from its own experiences, a model can refine its capabilities beyond human-labeled data. This framework opens avenues for richer multimodal interactions and cross-domain adaptation, pointing toward intelligence systems that can continually evolve autonomously. Our results suggest a promising path toward truly autonomous vision-language systems that improve themselves over time.

{
    \small
    \bibliographystyle{ieeenat_fullname}
    \bibliography{main}

@String(CVPR= {IEEE Conf. Comput. Vis. Pattern Recog.})

@String(CVPR  = {CVPR})

@article{vision-SR,
  title={Self-rewarding vision-language model via reasoning decomposition},
  author={Li, Zongxia and Yu, Wenhao and Huang, Chengsong and Liu, Rui and Liang, Zhenwen and Liu, Fuxiao and Che, Jingxi and Yu, Dian and Boyd-Graber, Jordan and Mi, Haitao and others},
  journal={arXiv preprint arXiv:2508.19652},
  year={2025}
}

@article{R-zero,
      title={R-Zero: Self-Evolving Reasoning LLM from Zero Data}, 
      author={Chengsong Huang and Wenhao Yu and Xiaoyang Wang and Hongming Zhang and Zongxia Li and Ruosen Li and Jiaxin Huang and Haitao Mi and Dong Yu},
      year={2025},
      eprint={2508.05004},
      archivePrefix={arXiv},
      primaryClass={cs.LG},
      url={https://arxiv.org/abs/2508.05004}, 
}

@misc{xia2025visionaryr1mitigatingshortcutsvisual,
      title={Visionary-R1: Mitigating Shortcuts in Visual Reasoning with Reinforcement Learning}, 
      author={Jiaer Xia and Yuhang Zang and Peng Gao and Yixuan Li and Kaiyang Zhou},
      year={2025},
      eprint={2505.14677},
      archivePrefix={arXiv},
      primaryClass={cs.CV},
      url={https://arxiv.org/abs/2505.14677}, 
}

@misc{feng2025videor1reinforcingvideoreasoning,
      title={Video-R1: Reinforcing Video Reasoning in MLLMs}, 
      author={Kaituo Feng and Kaixiong Gong and Bohao Li and Zonghao Guo and Yibing Wang and Tianshuo Peng and Junfei Wu and Xiaoying Zhang and Benyou Wang and Xiangyu Yue},
      year={2025},
      eprint={2503.21776},
      archivePrefix={arXiv},
      primaryClass={cs.CV},
      url={https://arxiv.org/abs/2503.21776}, 
}

@misc{chu2024visionllamaunifiedllamabackbone,
      title={VisionLLaMA: A Unified LLaMA Backbone for Vision Tasks}, 
      author={Xiangxiang Chu and Jianlin Su and Bo Zhang and Chunhua Shen},
      year={2024},
      eprint={2403.00522},
      archivePrefix={arXiv},
      primaryClass={cs.CV},
      url={https://arxiv.org/abs/2403.00522}, 
}

@misc{bai2023qwenvlversatilevisionlanguagemodel,
      title={Qwen-VL: A Versatile Vision-Language Model for Understanding, Localization, Text Reading, and Beyond}, 
      author={Jinze Bai and Shuai Bai and Shusheng Yang and Shijie Wang and Sinan Tan and Peng Wang and Junyang Lin and Chang Zhou and Jingren Zhou},
      year={2023},
      eprint={2308.12966},
      archivePrefix={arXiv},
      primaryClass={cs.CV},
      url={https://arxiv.org/abs/2308.12966}, 
}

@misc{li2022blipbootstrappinglanguageimagepretraining,
      title={BLIP: Bootstrapping Language-Image Pre-training for Unified Vision-Language Understanding and Generation}, 
      author={Junnan Li and Dongxu Li and Caiming Xiong and Steven Hoi},
      year={2022},
      eprint={2201.12086},
      archivePrefix={arXiv},
      primaryClass={cs.CV},
      url={https://arxiv.org/abs/2201.12086}, 
}

@misc{liu2023visualinstructiontuning,
      title={Visual Instruction Tuning}, 
      author={Haotian Liu and Chunyuan Li and Qingyang Wu and Yong Jae Lee},
      year={2023},
      eprint={2304.08485},
      archivePrefix={arXiv},
      primaryClass={cs.CV},
      url={https://arxiv.org/abs/2304.08485}, 
}

@article{absolute,
  title={Absolute zero: Reinforced self-play reasoning with zero data},
  author={Zhao, Andrew and Wu, Yiran and Yue, Yang and Wu, Tong and Xu, Quentin and Lin, Matthieu and Wang, Shenzhi and Wu, Qingyun and Zheng, Zilong and Huang, Gao},
  journal={arXiv preprint arXiv:2505.03335},
  year={2025}
}

@article{grpo,
  title={Deepseekmath: Pushing the limits of mathematical reasoning in open language models},
  author={Shao, Zhihong and Wang, Peiyi and Zhu, Qihao and Xu, Runxin and Song, Junxiao and Bi, Xiao and Zhang, Haowei and Zhang, Mingchuan and Li, YK and Wu, Yang and others},
  journal={arXiv preprint arXiv:2402.03300},
  year={2024}
}

@article{rlvr,
  title={Tulu 3: Pushing frontiers in open language model post-training},
  author={Lambert, Nathan and Morrison, Jacob and Pyatkin, Valentina and Huang, Shengyi and Ivison, Hamish and Brahman, Faeze and Miranda, Lester James V and Liu, Alisa and Dziri, Nouha and Lyu, Shane and others},
  journal={arXiv preprint arXiv:2411.15124},
  year={2024}
}

@article{Mm-vet,
  title={Mm-vet: Evaluating large multimodal models for integrated capabilities},
  author={Yu, Weihao and Yang, Zhengyuan and Li, Linjie and Wang, Jianfeng and Lin, Kevin and Liu, Zicheng and Wang, Xinchao and Wang, Lijuan},
  journal={arXiv preprint arXiv:2308.02490},
  year={2023}
}

@inproceedings{mmmu,
  title={Mmmu: A massive multi-discipline multimodal understanding and reasoning benchmark for expert agi},
  author={Yue, Xiang and Ni, Yuansheng and Zhang, Kai and Zheng, Tianyu and Liu, Ruoqi and Zhang, Ge and Stevens, Samuel and Jiang, Dongfu and Ren, Weiming and Sun, Yuxuan and others},
  booktitle={Proceedings of the IEEE/CVF Conference on Computer Vision and Pattern Recognition},
  pages={9556--9567},
  year={2024}
}

@article{VisNumBench,
  title={VisNumBench: Evaluating Number Sense of Multimodal Large Language Models},
  author={Weng, Tengjin and Wang, Jingyi and Jiang, Wenhao and Ming, Zhong},
  journal={arXiv preprint arXiv:2503.14939},
  year={2025}
}

@InProceedings{Li_2025_CVPR,
    author    = {Li, Zongxia and Wu, Xiyang and Du, Hongyang and Liu, Fuxiao and Nghiem, Huy and Shi, Guangyao},
    title     = {A Survey of State of the Art Large Vision Language Models: Benchmark Evaluations and Challenges},
    booktitle = {Proceedings of the Computer Vision and Pattern Recognition Conference (CVPR) Workshops},
    month     = {June},
    year      = {2025},
    pages     = {1587-1606}
}

@InProceedings{Guan_2024_CVPR,
    author    = {Guan, Tianrui and Liu, Fuxiao and Wu, Xiyang and Xian, Ruiqi and Li, Zongxia and Liu, Xiaoyu and Wang, Xijun and Chen, Lichang and Huang, Furong and Yacoob, Yaser and Manocha, Dinesh and Zhou, Tianyi},
    title     = {HallusionBench: An Advanced Diagnostic Suite for Entangled Language Hallucination and Visual Illusion in Large Vision-Language Models},
    booktitle = {Proceedings of the IEEE/CVF Conference on Computer Vision and Pattern Recognition (CVPR)},
    month     = {June},
    year      = {2024},
    pages     = {14375-14385}
}

@misc{li2025videohalluevaluatingmitigatingmultimodal,
      title={VideoHallu: Evaluating and Mitigating Multi-modal Hallucinations on Synthetic Video Understanding}, 
      author={Zongxia Li and Xiyang Wu and Guangyao Shi and Yubin Qin and Hongyang Du and Tianyi Zhou and Dinesh Manocha and Jordan Lee Boyd-Graber},
      year={2025},
      eprint={2505.01481},
      archivePrefix={arXiv},
      primaryClass={cs.CV},
      url={https://arxiv.org/abs/2505.01481}, 
}

@misc{fan2025sophiavlr1reinforcingmllmsreasoning,
      title={SophiaVL-R1: Reinforcing MLLMs Reasoning with Thinking Reward}, 
      author={Kaixuan Fan and Kaituo Feng and Haoming Lyu and Dongzhan Zhou and Xiangyu Yue},
      year={2025},
      eprint={2505.17018},
      archivePrefix={arXiv},
      primaryClass={cs.CV},
      url={https://arxiv.org/abs/2505.17018}, 
}

@misc{wang2025visionzeroscalablevlmselfimprovement,
      title={Vision-Zero: Scalable VLM Self-Improvement via Strategic Gamified Self-Play}, 
      author={Qinsi Wang and Bo Liu and Tianyi Zhou and Jing Shi and Yueqian Lin and Yiran Chen and Hai Helen Li and Kun Wan and Wentian Zhao},
      year={2025},
      eprint={2509.25541},
      archivePrefix={arXiv},
      primaryClass={cs.CV},
      url={https://arxiv.org/abs/2509.25541}, 
}

@misc{bordes2024introductionvisionlanguagemodeling,
      title={An Introduction to Vision-Language Modeling}, 
      author={Florian Bordes and Richard Yuanzhe Pang and Anurag Ajay and Alexander C. Li and Adrien Bardes and Suzanne Petryk and Oscar Mañas and Zhiqiu Lin and Anas Mahmoud and Bargav Jayaraman and Mark Ibrahim and Melissa Hall and Yunyang Xiong and Jonathan Lebensold and Candace Ross and Srihari Jayakumar and Chuan Guo and Diane Bouchacourt and Haider Al-Tahan and Karthik Padthe and Vasu Sharma and Hu Xu and Xiaoqing Ellen Tan and Megan Richards and Samuel Lavoie and Pietro Astolfi and Reyhane Askari Hemmat and Jun Chen and Kushal Tirumala and Rim Assouel and Mazda Moayeri and Arjang Talattof and Kamalika Chaudhuri and Zechun Liu and Xilun Chen and Quentin Garrido and Karen Ullrich and Aishwarya Agrawal and Kate Saenko and Asli Celikyilmaz and Vikas Chandra},
      year={2024},
      eprint={2405.17247},
      archivePrefix={arXiv},
      primaryClass={cs.LG},
      url={https://arxiv.org/abs/2405.17247}, 
}

@misc{sehgal2025selfevolvingvisualconceptlibrary,
      title={Self-Evolving Visual Concept Library using Vision-Language Critics}, 
      author={Atharva Sehgal and Patrick Yuan and Ziniu Hu and Yisong Yue and Jennifer J. Sun and Swarat Chaudhuri},
      year={2025},
      eprint={2504.00185},
      archivePrefix={arXiv},
      primaryClass={cs.CV},
      url={https://arxiv.org/abs/2504.00185}, 
}

@misc{chen2025c2evocoevolvingmultimodaldata,
      title={C2-Evo: Co-Evolving Multimodal Data and Model for Self-Improving Reasoning}, 
      author={Xiuwei Chen and Wentao Hu and Hanhui Li and Jun Zhou and Zisheng Chen and Meng Cao and Yihan Zeng and Kui Zhang and Yu-Jie Yuan and Jianhua Han and Hang Xu and Xiaodan Liang},
      year={2025},
      eprint={2507.16518},
      archivePrefix={arXiv},
      primaryClass={cs.CV},
      url={https://arxiv.org/abs/2507.16518}, 
}

@misc{gu2025surveyllmasajudge,
      title={A Survey on LLM-as-a-Judge}, 
      author={Jiawei Gu and Xuhui Jiang and Zhichao Shi and Hexiang Tan and Xuehao Zhai and Chengjin Xu and Wei Li and Yinghan Shen and Shengjie Ma and Honghao Liu and Saizhuo Wang and Kun Zhang and Yuanzhuo Wang and Wen Gao and Lionel Ni and Jian Guo},
      year={2025},
      eprint={2411.15594},
      archivePrefix={arXiv},
      primaryClass={cs.CL},
      url={https://arxiv.org/abs/2411.15594}, 
}

@inproceedings{li-etal-2025-llm-judge,
    title = "{LLM}-as-a-Judge Failures at Automating the Identification of Poor Quality Outputs in Free-Form Texts",
    author = "Li, Zongxia  and
      Wu, Xiyang  and
      Mondal, Ishani  and
      Siu, Alexa  and
      Boyd-Graber, Jordan Lee  and
      Nenkova, Ani",
    editor = "Dong, Yue  and
      Xiao, Wen  and
      Zhang, Haopeng  and
      Zhang, Rui  and
      Ernst, Ori  and
      Wang, Lu  and
      Liu, Fei",
    booktitle = "Proceedings of The 5th New Frontiers in Summarization Workshop",
    month = nov,
    year = "2025",
    address = "Hybrid",
    publisher = "Association for Computational Linguistics",
    url = "https://aclanthology.org/2025.newsum-main.1/",
    doi = "10.18653/v1/2025.newsum-main.1",
    pages = "1--16",
    ISBN = "979-8-89176-337-1"
}

@misc{tong2025gamerlsynthesizingmultimodalverifiable,
      title={Game-RL: Synthesizing Multimodal Verifiable Game Data to Boost VLMs' General Reasoning}, 
      author={Jingqi Tong and Jixin Tang and Hangcheng Li and Yurong Mou and Ming Zhang and Jun Zhao and Yanbo Wen and Fan Song and Jiahao Zhan and Yuyang Lu and Chaoran Tao and Zhiyuan Guo and Jizhou Yu and Tianhao Cheng and Zhiheng Xi and Changhao Jiang and Zhangyue Yin and Yining Zheng and Weifeng Ge and Guanhua Chen and Tao Gui and Xipeng Qiu and Qi Zhang and Xuanjing Huang},
      year={2025},
      eprint={2505.13886},
      archivePrefix={arXiv},
      primaryClass={cs.CL},
      url={https://arxiv.org/abs/2505.13886}, 
}

@misc{zhang2025r1vllearningreasonmultimodal,
      title={R1-VL: Learning to Reason with Multimodal Large Language Models via Step-wise Group Relative Policy Optimization}, 
      author={Jingyi Zhang and Jiaxing Huang and Huanjin Yao and Shunyu Liu and Xikun Zhang and Shijian Lu and Dacheng Tao},
      year={2025},
      eprint={2503.12937},
      archivePrefix={arXiv},
      primaryClass={cs.AI},
      url={https://arxiv.org/abs/2503.12937}, 
}

@misc{jia2024describethenreasonimprovingmultimodalmathematical,
      title={Describe-then-Reason: Improving Multimodal Mathematical Reasoning through Visual Comprehension Training}, 
      author={Mengzhao Jia and Zhihan Zhang and Wenhao Yu and Fangkai Jiao and Meng Jiang},
      year={2024},
      eprint={2404.14604},
      archivePrefix={arXiv},
      primaryClass={cs.CL},
      url={https://arxiv.org/abs/2404.14604}, 
}

@misc{RealWorldQA2024,
  title        = {RealWorldQA: Real-World Spatial Understanding Benchmark},
  author       = {{xAI}},
  year         = {2024},
  howpublished = {\url{https://x.ai/blog/grok-1.5v-and-realworldqa}},
  note         = {CC BY-ND 4.0 license. Benchmark dataset released with Grok-1.5 Vision.}
}

@misc{zhang2024mathversedoesmultimodalllm,
      title={MathVerse: Does Your Multi-modal LLM Truly See the Diagrams in Visual Math Problems?}, 
      author={Renrui Zhang and Dongzhi Jiang and Yichi Zhang and Haokun Lin and Ziyu Guo and Pengshuo Qiu and Aojun Zhou and Pan Lu and Kai-Wei Chang and Peng Gao and Hongsheng Li},
      year={2024},
      eprint={2403.14624},
      archivePrefix={arXiv},
      primaryClass={cs.CV},
      url={https://arxiv.org/abs/2403.14624}, 
}

@misc{wang2024measuringmultimodalmathematicalreasoning,
      title={Measuring Multimodal Mathematical Reasoning with MATH-Vision Dataset}, 
      author={Ke Wang and Junting Pan and Weikang Shi and Zimu Lu and Mingjie Zhan and Hongsheng Li},
      year={2024},
      eprint={2402.14804},
      archivePrefix={arXiv},
      primaryClass={cs.CV},
      url={https://arxiv.org/abs/2402.14804}, 
}

@misc{deepseekai2025deepseekr1incentivizingreasoningcapability,
      title={DeepSeek-R1: Incentivizing Reasoning Capability in LLMs via Reinforcement Learning}, 
      author={DeepSeek-AI and Daya Guo and Dejian Yang and Haowei Zhang and Junxiao Song and Ruoyu Zhang and Runxin Xu and Qihao Zhu and Shirong Ma and Peiyi Wang and Xiao Bi and Xiaokang Zhang and Xingkai Yu and Yu Wu and Z. F. Wu and Zhibin Gou and Zhihong Shao and Zhuoshu Li and Ziyi Gao and Aixin Liu and Bing Xue and Bingxuan Wang and Bochao Wu and Bei Feng and Chengda Lu and Chenggang Zhao and Chengqi Deng and Chenyu Zhang and Chong Ruan and Damai Dai and Deli Chen and Dongjie Ji and Erhang Li and Fangyun Lin and Fucong Dai and Fuli Luo and Guangbo Hao and Guanting Chen and Guowei Li and H. Zhang and Han Bao and Hanwei Xu and Haocheng Wang and Honghui Ding and Huajian Xin and Huazuo Gao and Hui Qu and Hui Li and Jianzhong Guo and Jiashi Li and Jiawei Wang and Jingchang Chen and Jingyang Yuan and Junjie Qiu and Junlong Li and J. L. Cai and Jiaqi Ni and Jian Liang and Jin Chen and Kai Dong and Kai Hu and Kaige Gao and Kang Guan and Kexin Huang and Kuai Yu and Lean Wang and Lecong Zhang and Liang Zhao and Litong Wang and Liyue Zhang and Lei Xu and Leyi Xia and Mingchuan Zhang and Minghua Zhang and Minghui Tang and Meng Li and Miaojun Wang and Mingming Li and Ning Tian and Panpan Huang and Peng Zhang and Qiancheng Wang and Qinyu Chen and Qiushi Du and Ruiqi Ge and Ruisong Zhang and Ruizhe Pan and Runji Wang and R. J. Chen and R. L. Jin and Ruyi Chen and Shanghao Lu and Shangyan Zhou and Shanhuang Chen and Shengfeng Ye and Shiyu Wang and Shuiping Yu and Shunfeng Zhou and Shuting Pan and S. S. Li and Shuang Zhou and Shaoqing Wu and Shengfeng Ye and Tao Yun and Tian Pei and Tianyu Sun and T. Wang and Wangding Zeng and Wanjia Zhao and Wen Liu and Wenfeng Liang and Wenjun Gao and Wenqin Yu and Wentao Zhang and W. L. Xiao and Wei An and Xiaodong Liu and Xiaohan Wang and Xiaokang Chen and Xiaotao Nie and Xin Cheng and Xin Liu and Xin Xie and Xingchao Liu and Xinyu Yang and Xinyuan Li and Xuecheng Su and Xuheng Lin and X. Q. Li and Xiangyue Jin and Xiaojin Shen and Xiaosha Chen and Xiaowen Sun and Xiaoxiang Wang and Xinnan Song and Xinyi Zhou and Xianzu Wang and Xinxia Shan and Y. K. Li and Y. Q. Wang and Y. X. Wei and Yang Zhang and Yanhong Xu and Yao Li and Yao Zhao and Yaofeng Sun and Yaohui Wang and Yi Yu and Yichao Zhang and Yifan Shi and Yiliang Xiong and Ying He and Yishi Piao and Yisong Wang and Yixuan Tan and Yiyang Ma and Yiyuan Liu and Yongqiang Guo and Yuan Ou and Yuduan Wang and Yue Gong and Yuheng Zou and Yujia He and Yunfan Xiong and Yuxiang Luo and Yuxiang You and Yuxuan Liu and Yuyang Zhou and Y. X. Zhu and Yanhong Xu and Yanping Huang and Yaohui Li and Yi Zheng and Yuchen Zhu and Yunxian Ma and Ying Tang and Yukun Zha and Yuting Yan and Z. Z. Ren and Zehui Ren and Zhangli Sha and Zhe Fu and Zhean Xu and Zhenda Xie and Zhengyan Zhang and Zhewen Hao and Zhicheng Ma and Zhigang Yan and Zhiyu Wu and Zihui Gu and Zijia Zhu and Zijun Liu and Zilin Li and Ziwei Xie and Ziyang Song and Zizheng Pan and Zhen Huang and Zhipeng Xu and Zhongyu Zhang and Zhen Zhang},
      year={2025},
      eprint={2501.12948},
      archivePrefix={arXiv},
      primaryClass={cs.CL},
      url={https://arxiv.org/abs/2501.12948}, 
}

@misc{peng2025lmmr1empowering3blmms,
      title={LMM-R1: Empowering 3B LMMs with Strong Reasoning Abilities Through Two-Stage Rule-Based RL}, 
      author={Yingzhe Peng and Gongrui Zhang and Miaosen Zhang and Zhiyuan You and Jie Liu and Qipeng Zhu and Kai Yang and Xingzhong Xu and Xin Geng and Xu Yang},
      year={2025},
      eprint={2503.07536},
      archivePrefix={arXiv},
      primaryClass={cs.CL},
      url={https://arxiv.org/abs/2503.07536}, 
}

@article{huang2025efficient,
  title={Efficient test-time scaling via self-calibration},
  author={Huang, Chengsong and Huang, Langlin and Leng, Jixuan and Liu, Jiacheng and Huang, Jiaxin},
  journal={arXiv preprint arXiv:2503.00031},
  year={2025}
}

@misc{MiMo-VL,
      title={MiMo-VL Technical Report}, 
      author={LLM-Core-Team Xiaomi},
      year={2025},
      eprint={2506.03569},
      archivePrefix={arXiv},
      primaryClass={cs.CL},
      url={https://arxiv.org/abs/2506.03569}, 
}

@misc{qwen2.5-VL,
    title = {Qwen2.5-VL},
    url = {https://qwenlm.github.io/blog/qwen2.5-vl/},
    author = {Qwen Team},
    month = {January},
    year = {2025}
}

@article{liu2025spice,
  title={SPICE: Self-Play In Corpus Environments Improves Reasoning},
  author={Liu, Bo and Jin, Chuanyang and Kim, Seungone and Yuan, Weizhe and Zhao, Wenting and Kulikov, Ilia and Li, Xian and Sukhbaatar, Sainbayar and Lanchantin, Jack and Weston, Jason},
  journal={arXiv preprint arXiv:2510.24684},
  year={2025}
}

@article{chen2025multi,
  title={Multi-Agent Evolve: LLM Self-Improve through Co-evolution},
  author={Chen, Yixing and Wang, Yiding and Zhu, Siqi and Yu, Haofei and Feng, Tao and Zhang, Muhan and Patwary, Mostofa and You, Jiaxuan},
  journal={arXiv preprint arXiv:2510.23595},
  year={2025}
}

@article{lu2025search,
  title={Search Self-play: Pushing the Frontier of Agent Capability without Supervision},
  author={Lu, Hongliang and Wen, Yuhang and Cheng, Pengyu and Ding, Ruijin and Xu, Haotian and Guo, Jiaqi and Wang, Chutian and Chen, Haonan and Jiang, Xiaoxi and Jiang, Guanjun},
  journal={arXiv preprint arXiv:2510.18821},
  year={2025}
}

@article{liu2025deep,
  title={Deep Self-Evolving Reasoning},
  author={Liu, Zihan and Zheng, Shun and Wen, Xumeng and Wang, Yang and Bian, Jiang and Yang, Mao},
  journal={arXiv preprint arXiv:2510.17498},
  year={2025}
}

@article{xu2025genius,
  title={Genius: A generalizable and purely unsupervised self-training framework for advanced reasoning},
  author={Xu, Fangzhi and Yan, Hang and Ma, Chang and Zhao, Haiteng and Sun, Qiushi and Cheng, Kanzhi and He, Junxian and Liu, Jun and Wu, Zhiyong},
  journal={arXiv preprint arXiv:2504.08672},
  year={2025}
}

@article{kuba2025language,
  title={Language self-play for data-free training},
  author={Kuba, Jakub Grudzien and Gu, Mengting and Ma, Qi and Tian, Yuandong and Mohan, Vijai},
  journal={arXiv preprint arXiv:2509.07414},
  year={2025}
}

@article{yang2025spell,
  title={SPELL: Self-Play Reinforcement Learning for evolving Long-Context Language Models},
  author={Yang, Ziyi and Shen, Weizhou and Chen, Ruijun and Li, Chenliang and Wan, Fanqi and Yan, Ming and Quan, Xiaojun and Huang, Fei},
  journal={arXiv preprint arXiv:2509.23863},
  year={2025}
}

@article{wang2025socratic,
  title={Socratic-Zero: Bootstrapping Reasoning via Data-Free Agent Co-evolution},
  author={Wang, Shaobo and Jiao, Zhengbo and Zhang, Zifan and Peng, Yilang and Ze, Xu and Yang, Boyu and Wang, Wei and Wei, Hu and Zhang, Linfeng},
  journal={arXiv preprint arXiv:2509.24726},
  year={2025}
}

@incollection{schmidhuber2007godel,
  title={G{\"o}del machines: Fully self-referential optimal universal self-improvers},
  author={Schmidhuber, J{\"u}rgen},
  booktitle={Artificial general intelligence},
  pages={199--226},
  year={2007},
  publisher={Springer}
}

@article{clune2019ai,
  title={AI-GAs: AI-generating algorithms, an alternate paradigm for producing general artificial intelligence},
  author={Clune, Jeff},
  journal={arXiv preprint arXiv:1905.10985},
  year={2019}
}

@article{yang2024weak,
  title={Weak-to-strong reasoning},
  author={Yang, Yuqing and Ma, Yan and Liu, Pengfei},
  journal={arXiv preprint arXiv:2407.13647},
  year={2024}
}

@article{dai2025cde,
  title={Cde: Curiosity-driven exploration for efficient reinforcement learning in large language models},
  author={Dai, Runpeng and Song, Linfeng and Liu, Haolin and Liang, Zhenwen and Yu, Dian and Mi, Haitao and Tu, Zhaopeng and Liu, Rui and Zheng, Tong and Zhu, Hongtu and others},
  journal={arXiv preprint arXiv:2509.09675},
  year={2025}
}

@article{zhou2025evolving,
  title={Evolving language models without labels: Majority drives selection, novelty promotes variation},
  author={Zhou, Yujun and Liang, Zhenwen and Liu, Haolin and Yu, Wenhao and Panaganti, Kishan and Song, Linfeng and Yu, Dian and Zhang, Xiangliang and Mi, Haitao and Yu, Dong},
  journal={arXiv preprint arXiv:2509.15194},
  year={2025}
}

@article{liu2025vogue,
  title={Vogue: Guiding exploration with visual uncertainty improves multimodal reasoning},
  author={Liu, Rui and Yu, Dian and Zheng, Tong and Dai, Runpeng and Li, Zongxia and Yu, Wenhao and Liang, Zhenwen and Song, Linfeng and Mi, Haitao and Tokekar, Pratap and others},
  journal={arXiv preprint arXiv:2510.01444},
  year={2025}
}

@article{chen2025pass,
  title={Pass@ k training for adaptively balancing exploration and exploitation of large reasoning models},
  author={Chen, Zhipeng and Qin, Xiaobo and Wu, Youbin and Ling, Yue and Ye, Qinghao and Zhao, Wayne Xin and Shi, Guang},
  journal={arXiv preprint arXiv:2508.10751},
  year={2025}
}

@article{cheng2025reasoning,
  title={Reasoning with exploration: An entropy perspective},
  author={Cheng, Daixuan and Huang, Shaohan and Zhu, Xuekai and Dai, Bo and Zhao, Wayne Xin and Zhang, Zhenliang and Wei, Furu},
  journal={arXiv preprint arXiv:2506.14758},
  year={2025}
}

@article{zheng2025parallel,
  title={Parallel-r1: Towards parallel thinking via reinforcement learning},
  author={Zheng, Tong and Zhang, Hongming and Yu, Wenhao and Wang, Xiaoyang and Dai, Runpeng and Liu, Rui and Bao, Huiwen and Huang, Chengsong and Huang, Heng and Yu, Dong},
  journal={arXiv preprint arXiv:2509.07980},
  year={2025}
}

@article{jin2025search,
  title={Search-r1: Training llms to reason and leverage search engines with reinforcement learning},
  author={Jin, Bowen and Zeng, Hansi and Yue, Zhenrui and Yoon, Jinsung and Arik, Sercan and Wang, Dong and Zamani, Hamed and Han, Jiawei},
  journal={arXiv preprint arXiv:2503.09516},
  year={2025}
}

@article{zhao2025one,
  title={One token to fool llm-as-a-judge},
  author={Zhao, Yulai and Liu, Haolin and Yu, Dian and Kung, Sunyuan and Chen, Meijia and Mi, Haitao and Yu, Dong},
  journal={arXiv preprint arXiv:2507.08794},
  year={2025}
}
}

\clearpage
\setcounter{page}{1}
\maketitlesupplementary

\section*{A.1 Detailed Training Dataset and Benchmarks}
\label{sec:rationale}
\paragraph{Training Dataset}
We use the image data from Vision-47K dataset~\cite{vision-SR, feng2025videor1reinforcingvideoreasoning}, which contains 47,000 web-sourced images covering a wide variety of domains. The dataset includes charts, medical images, educational exams, textbook illustrations, and driving simulation frames. For our purposes, we exclusively use the images themselves, omitting any associated questions and answers. The dataset consists of approximately 10K charts, 8K medical images, 12K educational images (from exams and textbooks), 7K driving scenes, and 10K miscellaneous images from various domains. All images were standardized to a resolution of 224$\times$224 pixels for model training.

\paragraph{Backbone Models and Training}
We trained three backbone models using \method{}:
\begin{itemize}
    \item \textbf{Qwen2.5-VL-3B-Instruct}\footnote{https://huggingface.co/Qwen/Qwen2.5-VL-3B-Instruct}: 3 billion parameters, fine-tuned with multimodal instruction data to enhance reasoning over visual-text tasks.
    \item \textbf{Qwen2.5-VL-7B-Instruct}\footnote{https://huggingface.co/Qwen/Qwen2.5-VL-7B-Instruct}: 7 billion parameters, trained under the same protocol with extended batch sizes and longer training schedules to improve complex reasoning and generalization.
    \item \textbf{Mimo-VL-7B-SFT}\footnote{https://huggingface.co/XiaomiMiMo/MiMo-VL-7B-SFT}: 7 billion parameters, optimized with supervised fine-tuning on multimodal datasets for better alignment with human instructions.
\end{itemize}

\paragraph{General Visual Understanding}
Four established benchmarks are used:
\begin{itemize}
    \item \textbf{MM-Vet}~\cite{Mm-vet}: Evaluates recognition, OCR, and visual math abilities using a unified LLM-based scoring metric. The dataset contains over 5,000 test samples with detailed scoring for each subtask.
    \item \textbf{MMMU}~\cite{mmmu}: Cross-modal reasoning benchmark with 11.5K college-level multiple-choice questions spanning six academic disciplines. Each question is image-based and designed to test subject knowledge and reasoning ability.
    \item \textbf{RealWorldQA}~\cite{RealWorldQA2024}: Contains approximately 700 real-world images paired with spatially grounded questions. Evaluation emphasizes spatial reasoning and contextual understanding.
    \item \textbf{VisNumBench}~\cite{VisNumBench}: Focused on visual number sense, includes roughly 1.9K questions involving numerical attributes, comparisons, and estimations.
\end{itemize}

\paragraph{Multimodal Mathematical Reasoning}
Two specialized benchmarks:
\begin{itemize}
    \item \textbf{MathVerse}~\cite{zhang2024mathversedoesmultimodalllm}: 2.6K diagram-centric questions covering geometry, functions, and algebra, provided in multiple visual-text formats.
    \item \textbf{MATH-Vision}~\cite{wang2024measuringmultimodalmathematicalreasoning}: Approximately 3K competition-level problems across 16 subjects and five difficulty tiers. Focuses on integrating visual information into advanced mathematical reasoning.
\end{itemize}

\paragraph{Visual Hallucination Detection}
\textbf{HallusionBench}~\cite{Guan_2024_CVPR} is used to identify model errors caused by either language-only hallucinations or visual illusions. Evaluation is conducted in a simple yes/no format, enabling precise measurement of hallucination rates and error types.

\section*{A.2 Training Configuration}
\paragraph{Image-based Questioner Configuration.}
The Questioner is trained using a vision–language model with a maximum context window of 8192 tokens. 
The training set consists of 47K multimodal samples (Vision-SR1-47K), and evaluation is performed on the MMStar benchmark. 
Each sample uses \texttt{problem}, \texttt{answer}, and \texttt{images} as the prompt, label, and image fields, respectively. 
During rollouts, the Questioner generates 8 candidate questions per input. 
For the 3B model, we train for 20 steps, and for the 7B model, we train for 10 steps. In this setup, four GPUs run the vLLM service to provide reward signals, while the other four GPUs are used for training.
The model parameters are loaded from a specified Questioner checkpoint, and all checkpoints are saved under the designated experiment directory. 
Validation before training is disabled to maximize efficiency during early-stage learning.

\begin{figure*}[h!] % * 表示跨双栏
\begin{tcolorbox}[
  colback=blue!5!white,
  colframe=blue!60!black,
  title=\textbf{\concept{} Prompt Template},
  fonttitle=\bfseries,
  sharp corners,
  width=\textwidth  % 确保宽度占满双栏
]

You are an intelligent Question Generator. Your task is to create a question based on the given image.  
\par
Requirements (must follow exactly):  

1. Analyze the image carefully and understand all details.  
2. Generate exactly one question that is directly related to the image.  
3. Choose the question type from only one of the following:  
- multiple choice (Yes/No or four options labeled A, B, C, D; only one correct answer)  
- numerical (requires a specific numeric answer)  
- regression (requires predicting a continuous value, such as a measurement, quantity, or coordinate)  
4. The question must require analysis or reasoning, not just description.  
5. Output must be strictly in format $<question>X</question>$, with nothing else:
% \\
% \par
% \vspace{0.5em}
% The following block:   
% \par
% $<question>X</question>$

Strict rules:  
- Do not use any other labels, punctuation, or formatting.  
- Do not add commentary, explanations, or extra text.  
Example of correct output:   
\par
$<question>How \ \ many \ \ clubs \ \ are \ \ there</question>$
\end{tcolorbox}
\end{figure*}

\begin{figure*}[h!]
\begin{tcolorbox}[
  colback=blue!5!white,   % soft yellow background
  colframe=blue!60!black,  % warm golden frame
  title=\textbf{\textit{Multimodal Reasoner} Prompt Template},
  fonttitle=\bfseries,
  sharp corners,
  width=\textwidth
]
Please reason step by step carefully based on the question:  + content +  and the image.  After completing your reasoning, you MUST output the final, clean, and concise answer strictly inside  + \textbackslash boxed\{\} + .  The final answer MUST appear inside \textbackslash boxed\{\}, and nowhere else.  If there is no boxed answer, your response is considered incorrect. 
\end{tcolorbox}
\end{figure*}

\begin{figure*}[h!]
\begin{tcolorbox}[
  colback=blue!5!white,   % soft yellow background
  colframe=blue!60!black,  % warm golden frame
  title=\textbf{\textit{LLM-as-Judge} Prompt Template},
  fonttitle=\bfseries,
  sharp corners,
  width=\textwidth
]
You are an answer evaluation assistant. Your task is to judge whether two answers are substantially equivalent. When evaluating, you should ignore superficial differences such as format, spaces, punctuation, case, etc., and focus on whether they are consistent in core content, logical meaning and information expression. The judgment criteria should be lenient and inclusive, as long as the expressed meaning is basically the same, it is considered equivalent.
\end{tcolorbox}
\end{figure*}

\paragraph{Multimodal Reasoner Configuration.}
The Solver is trained using chain-of-thought reinforcement learning. 
Its output length is capped at 4096 tokens, and prompts are constructed using a dedicated Jinja template to enforce a consistent reasoning format. 
Training uses the self-play dataset produced by the Questioner, while evaluation again uses MMStar. 
To ensure stable learning under long sequences, we adopt a conservative micro-batch size of~1 for both updates and experience rollouts. 
The rollout engine supports up to 20K batched tokens per forward pass. 
The number of training steps for the Solver is set to be the same as for the Questioner.

\section*{A.3 Prompt Templates}
The following three prompt templates define the core interaction structure used in our self-evolving Vision-Language Model. In this setup, multiple specialized roles—such as a question generator, a multimodal reasoner, and an evaluator—are orchestrated to form an autonomous learning loop. Each template specifies precise behavioral constraints that allow the model to generate tasks, solve them with step-by-step reasoning, and assess answer consistency. Together, these components establish a controlled self-play environment that enables the model to iteratively refine its reasoning capabilities without relying on external supervision.

\end{document}